\documentclass[journal]{IEEEtran}
\usepackage{amsmath,amsfonts}
\usepackage{algorithmic}
\usepackage{algorithm}
\usepackage{array}
\usepackage[caption=false,font=normalsize,labelfont=sf,textfont=sf]{subfig}
\usepackage{textcomp}
\usepackage{stfloats}
\usepackage{url}
\usepackage{verbatim}
\usepackage{graphicx}
\usepackage{multirow}
\usepackage{cite}
\usepackage{orcidlink}
\usepackage{tabularx}  
 \usepackage{booktabs}
\usepackage{hyperref}
\usepackage{newtxtext,newtxmath}
\hypersetup{
	colorlinks=true,
	linkcolor=blue,
	filecolor=magenta,
	urlcolor=magenta,
}
\begin{document}

\title{Feature Engineering is Not Dead: Reviving Classical Machine Learning with Entropy, HOG, and LBP Feature Fusion for Image Classification}


\author{Abhijit Sen\orcidlink{0000-0003-2783-1763},
Giridas Maiti\orcidlink{0000-0002-7813-6480}, 
Bikram K. Parida\orcidlink{0000-0003-1204-357X}, 
Bhanu P. Mishra,
Mahima Arya,  
Denys I. Bondar\orcidlink{0000-0002-3626-4804}

\thanks{A Sen, B.K. Parida and D.I. Bondar are with  Department of Physics and Engineering Physics, Tulane University, New Orleans, Louisiana 70118, USA. (email: asen1@tulane.edu; parida.bikram90.bkp@gmail.com; dbondar@tulane.edu).}
\thanks{G. Maiti is with Institute of Applied Geosciences, Karlsruhe Institute of Technology (KIT), Karlsruhe 76131, Germany. (email: giridas.maiti@kit.edu)}%
\thanks{B.P. Mishra is with Cure and Care Wellness Pvt Ltd, Bengaluru 560068, India.(email: pratapbhanumishra@gmail.com)}%
\thanks{M. Arya is with Amritha Vidya Vishwapeetham, Tamil Nadu 641112, India. (email: aryamahima@gmail.com)}%
\thanks{This work has been supported by Army Research Office (ARO) (grant W911NF-23-1-0288; program manager Dr.~James Joseph). The views and conclusions contained in this document are those of the authors and should not be interpreted as representing the official policies, either expressed or implied, of ARO or the U.S. Government. The U.S. Government is authorized to reproduce and distribute reprints for Government purposes notwithstanding any copyright notation herein. }
\thanks{Manuscript received July 20, 2025; revised August 8, 2025. (Corresponding author : Giridas Maiti)}%
\thanks{For those interested, the Github link of the implemented codes are available at \href{https://github.com/Generative-ML-for-Quantum-System/PE_based_image_classification}{https://github.com/Generative-ML-for-Quantum-System/PE\_based\_image\_classification}.}}

\markboth{Journal of \LaTeX\ Class Files,~Vol.~14, No.~8, August~2021}%
{Shell \MakeLowercase{\textit{et al.}}: A Sample Article Using IEEEtran.cls for IEEE Journals}


\maketitle

\begin{abstract}
Feature engineering continues to play a critical role in image classification, particularly when interpretability and computational efficiency are prioritized over deep learning models with millions of parameters. In this study, we revisit classical machine learning based image classification through a novel approach centered on Permutation Entropy (PE), a robust and computationally lightweight measure traditionally used in time series analysis but rarely applied to image data. We extend PE to two-dimensional images and propose a multiscale, multi-orientation entropy-based feature extraction approach that characterizes spatial order and complexity along rows, columns, diagonals, anti-diagonals, and local patches of the image. To enhance the discriminatory power of the entropy features, we integrate two classic image descriptors: the Histogram of Oriented Gradients (HOG) to capture shape and edge structure, and Local Binary Patterns (LBP) to encode micro-texture of an image. The resulting hand-crafted feature set, comprising of 780 dimensions, is used to train Support Vector Machine (SVM) classifiers optimized through grid search. The proposed approach is evaluated on multiple benchmark datasets, including Fashion-MNIST, KMNIST, EMNIST, and CIFAR-10, where it delivers competitive classification performance without relying on deep architectures. Our results demonstrate that the fusion of PE with HOG and LBP provides a compact, interpretable, and effective alternative to computationally expensive and limited interpretable deep learning models. This shows a potential of entropy-based descriptors in image classification and contributes a lightweight and generalizable solution to interpretable machine learning in image classification and computer vision.

\end{abstract}


\section{Introduction}
\IEEEPARstart{P}{ermutation} entropy (PE), introduced by Bandt and Pompe~\cite{bandt2002permutation}, is a symbolic time-series complexity measure that captures the diversity of ordinal patterns in a signal. Unlike conventional entropy metrics such as Shannon or Rényi entropy~\cite{shannon1948mathematical, renyi1961measures}, which are derived from amplitude-based probability distributions, PE encodes local temporal dynamics by analyzing the relative ordering of values within sliding windows. Due to its robustness to noise, computational simplicity and model-free formulation, PE has found widespread application in physiological signal analysis, including EEG-based seizure detection~\cite{nicolaou2012use}, heart rate variability~\cite{zunino2012permutation}, and sleep stage classification~\cite{li2016automatic}, , and fMRI ~\cite{xiao2025mse_asd} analysis.

Despite its success in 1D temporal domains, permutation entropy (PE) remains largely unexplored in natural image classification. A few previous works have used entropy-related ideas in medical imaging—for example, entropy-guided patch sampling within CT liver segmentation pipelines~\cite{qin2018sbbscnn}—but these
approaches are task- and modality-specific and do not treat PE as a general-purpose image descriptor. Beyond medical image analysis, entropy has a long-standing role in classical image processing. Early work used
entropy principles for fundamental operations such as gray-level thresholding and segmentation~\cite{kapur1985_entropy_thresholding}. More recently, several two-dimensional ordinal/entropy variants
have been proposed specifically for texture characterization and discrimination
~\cite{gaudencio2022_pe2d_aape2d,zunino2016_ms_cecp_textures,morel2021_mpe2d}. However, to the best of our
knowledge, PE has not been employed as a primary, global handcrafted feature for natural image
classification within a unified multi-orientation, multi-scale design.

This gap is particularly notable given the computational cost and complexity associated with modern deep learning approaches. Convolutional neural networks (CNNs) such as VGG~\cite{simonyan2015vgg}, ResNet~\cite{he2016deep}, and MobileNet~\cite{howard2017mobilenets} dominate datasets like Fashion-MNIST~\cite{xiao2017fashion}, achieving 93--94\% classification accuracy. However, these models come at the expense of millions of trainable parameters, extensive training resources, and limited interpretability. For example, ResNet-18 contains more than 11 million parameters~\cite{he2016deep}, and even lightweight models such as MobileNet require substantial GPU memory and training time. Furthermore, the black-box nature of deep learning impedes transparency and scientific understanding.

In this work, we present a novel, interpretable, and computationally lightweight approach that uses PE for full-image classification. The PE features are combined with two complementary image descriptors, the Histogram of Oriented Gradients (HOG)~\cite{dalal2005} and the Local Binary Patterns (LBP)~\cite{ojala2002}. HOG captures shape and edge structure of an image, useful for detecting object outlines and silhouettes, whereas LBP captures micro-texture such as spots, flat areas, edges and corners, and good at capturing fine repetitive patterns. We combine PE, HOG and LBP features to form a compact, handcrafted feature vector of \textit{n-dimensions}. Let us clarify in what sense the handcrafted features are interpretable. By interpretable, we mean that each feature group has a clear, known semantic meaning. HOG captures oriented edges and shape, LBP encodes local texture patterns, and PE measures local complexity and irregularity. Unlike CNNs, where features are learned automatically and often require post-hoc tools (e.g., Grad-CAM, SHAP) to explain decisions, our descriptors are predefined and semantically grounded, allowing us to directly identify what image properties the classifier relies upon.

We use these handcrafted feature vectors and employ Support Vector Machines (SVM) \cite{cortes1995support}, a classical machine learning method, for image classification work across multiple benchmark datasets such as Fashion-MNIST, KMNIST, EMNIST, and CIFAR-10. In particular, we  achieve up to \textbf{91.23\% test accuracy} on Fashion-MNIST benchmark dataset. All previous test accuracy results~\cite{fashionmnistweb} in the Fashion-MNIST benchmark dataset, that used \texttt{scikit-learn} library and evaluated 129 classifiers with various parameter configurations (excluding deep learning methods), reported a maximum accuracy of \textbf{89.7\%} using SVM with parameters \texttt{C=10, kernel=poly}~\cite{fashionmnistweb}. In contrast, our method achieves an accuracy of 91.23\% for the first time using purely classical machine learning techniques applied on hand-crafted features extracted from the Fashion-MNIST dataset. We also apply this method on other benchmark datasets (KMNIST, EMNIST, and CIFAR-10) and show competitive performance.

The rest of the paper is organized as follows: In Section \ref{sec:PE}, we present a detailed review of permutation entropy; Section \ref{sec:HOG_LBP} discusses the classical image descriptors HOG and LBP employed in this work; Section \ref{sec:FE_extraction} describes our feature-extraction pipeline in detail; Section \ref{sec:SVM} briefly introduces classical SVM-based machine learning; Section \ref{sec:Dataset} explains our rationale for selecting the benchmark datasets and provides their descriptions; Section \ref{sec:Results} reports the results obtained using our novel feature-extraction pipeline and compares them with previous studies; finally, Section \ref{sec:Conclusion} concludes the paper and outlines future implications.

\section{ Permutation Entropy (PE)}\label{sec:PE}
\textbf{Shannon entropy} is a foundational concept in information theory that quantifies the average amount of uncertainty or information contained in a probability distribution. For a discrete random variable $X$ with possible outcomes $\{x_1, x_2, \dots, x_n\}$ and corresponding probabilities $\{p_1, p_2, \dots, p_n\}$, the Shannon entropy is defined as:

\begin{equation}
H(X) = -\sum_{i=1}^{n} p_i \log p_i.
\end{equation}

Entropy reaches its maximum when all outcomes are equally likely (i.e., the distribution is uniform), and is lower when the distribution is more predictable or concentrated.

\vspace{1em}
\textbf{Permutation Entropy} is a specialized form of entropy introduced by Bandt and Pompe~\cite{bandt2002permutation} to measure the complexity of time series data. Unlike Shannon entropy, which requires an explicit probability distribution over symbols or values, PE evaluates the entropy of the \textit{ordinal patterns} — that is, the relative ordering of values within segments of the time series.

\begin{figure}[htbp!]
\centering
\includegraphics[width= 3.8 in]{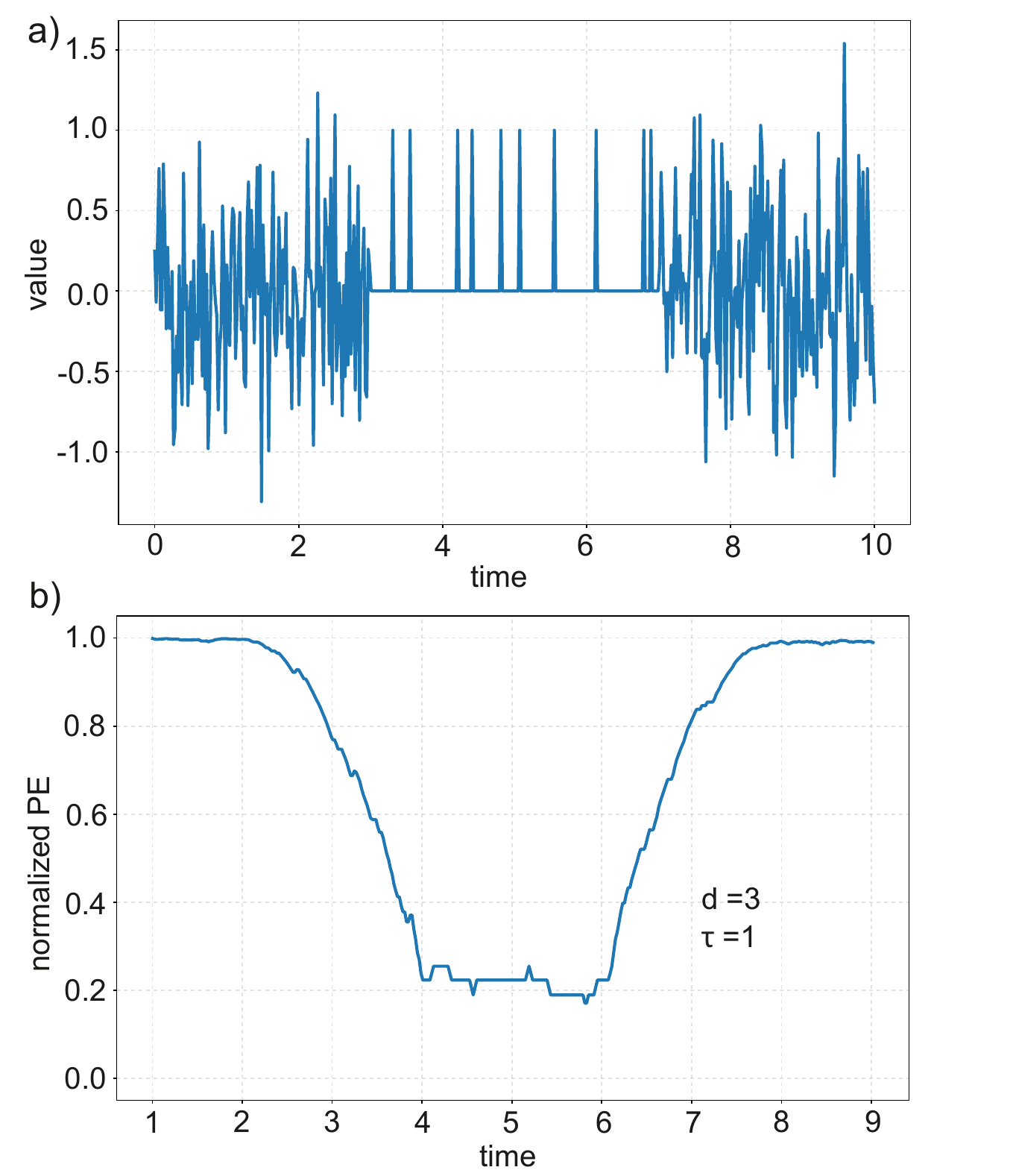}
\caption{Visualization of PE from a time-series data: (a) a sample time series data; and (b) corresponding normalized PE with embedding dimension \textit{d} = 3 and time delay $\tau $ =1. Note that in high irregularity domain PE is near 1 and in high regularity or predictable domain PE is near 0. }
\label{fig_1}
\end{figure}

To understand PE, let's consider a scalar time series $\{x_t\}_{t=1}^N$. To extract ordinal patterns, we first embed the series in a higher-dimensional space using parameters embedding dimension $d \geq 2$ (length of each pattern) and time delay $\tau \geq 1$ (step between points in a pattern). This results in a sequence of delay vectors:
\begin{align}
\mathbf{X}_t = (&x_t, x_{t+\tau}, x_{t+2\tau}, \dots, \nonumber\\
               &x_{t+(d-1)\tau}), \quad t = 1, 2, \dots, N - (d - 1)\tau.
\end{align}

For each vector $\mathbf{X}_t$, we determine its \textit{ordinal pattern} — the permutation $\pi \in S_d$ that sorts the elements in ascending order:
\[
x_{t+\pi_1\tau} \leq x_{t+\pi_2\tau} \leq \cdots \leq x_{t+\pi_d\tau}.
\]
If ties occur, they are resolved by the order of appearance. Next, we compute the relative frequency $P(\pi)$ of each permutation $\pi$ across the time series. The PE is then defined as the Shannon entropy of this distribution:
\begin{equation}
H_d = -\sum_{\pi \in S_d} P(\pi) \log P(\pi),
\end{equation}
where $S_d$ is the set of all $d!$ possible permutations of order $d$. To enable comparison across different embedding dimensions, it is often normalized:
\begin{equation}
h_d = \frac{H_d}{\log(d!)} \in [0, 1].
\end{equation}

For example, let us consider a short time series: $\{x_t\} = \{4, 7, 9, 10, 6\}$, with $d=3$ and $\tau=1$. The delay vectors are:
\begin{align*}
\mathbf{X}_1 &= (4, 7, 9) \rightarrow \pi = (1, 2, 3),\\
\mathbf{X}_2 &= (7, 9, 10) \rightarrow \pi = (1, 2, 3),\\
\mathbf{X}_3 &= (9, 10, 6) \rightarrow \pi = (3, 1, 2).
\end{align*}

From these, the ordinal patterns $(1,2,3)$ and $(3,1,2)$ occur with frequencies $2/3$ and $1/3$, respectively. The entropy is then:
\[
H_3 = -\left(\frac{2}{3} \log \frac{2}{3} + \frac{1}{3} \log \frac{1}{3}\right) \approx 0.918 \text{ bits},
\]
\[
h_3 = \frac{H_3}{\log(6)} \approx \frac{0.918}{\log(6)} \approx 0.51.
\]

A \textbf{low PE} (near 0) indicates high regularity or predictability—common in periodic or deterministic signals.
In contrasts a \textbf{high PE} (near 1) suggests randomness or high complexity—typical of stochastic or chaotic processes (Fig.~\ref{fig_1}).

In the context of image analysis, 2D grayscale image patches can be flattened into 1D sequences and analyzed using PE to quantify local structural complexity. This is particularly useful in medical imaging, texture classification, or anomaly detection, where small variations in spatial ordering can signal important features. Unlike pixel intensity-based methods, PE is sensitive to the shape and variation patterns, making it suitable for evaluating datasets with rich intra-class texture or fine-grained structural differences.

\section{Traditional Feature Descriptors: HOG and LBP}\label{sec:HOG_LBP}
\subsection{Histogram of Oriented Gradients (HOG)}
\begin{figure}[htbp!]
\centering
\includegraphics[width= 3.8 in]{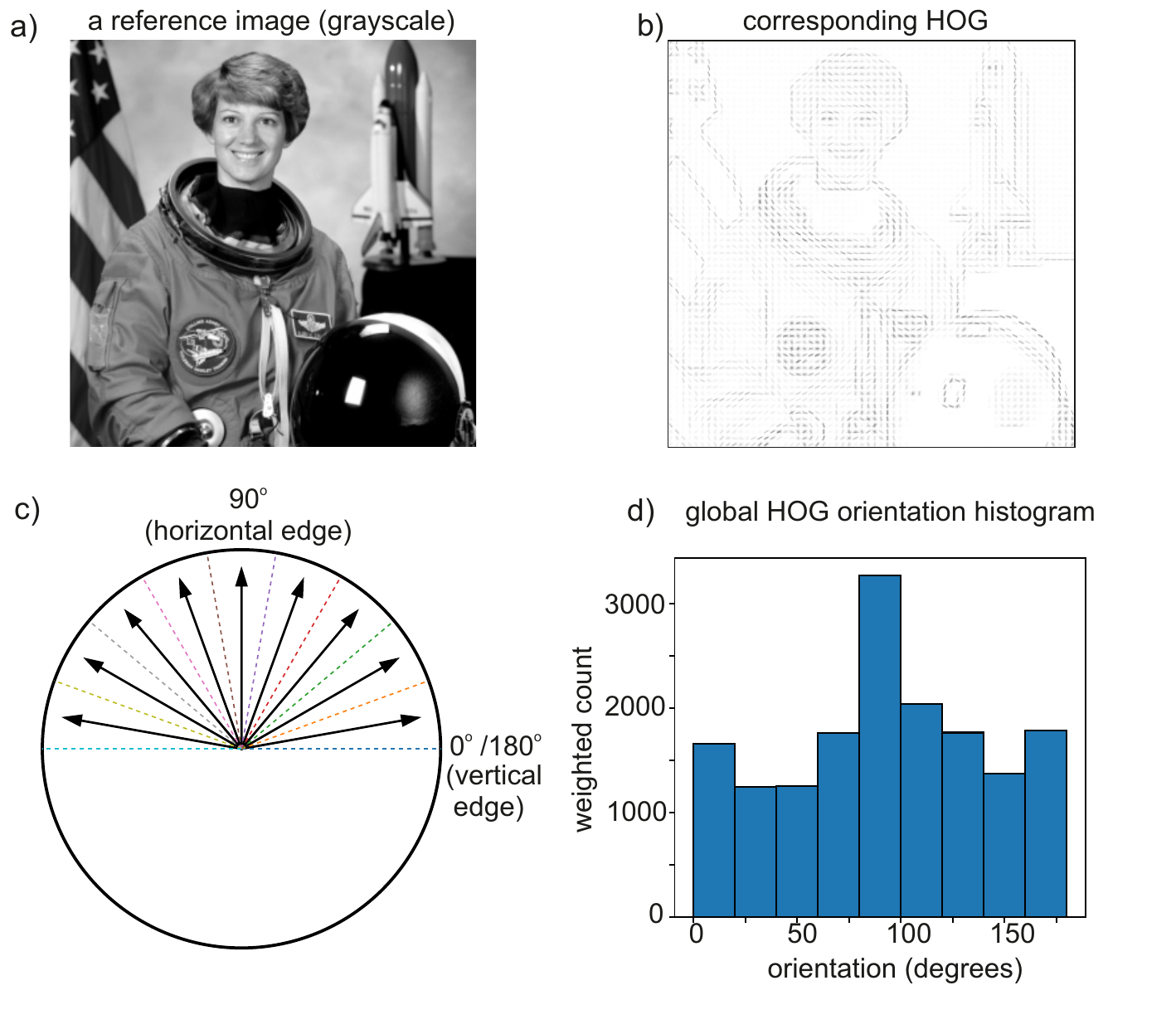}
\caption{Visualization of HOG from an image: (a) reference image (source: scikit-image); (b) corresponding HOG that nicely captures shape, outline and edges of the image; (c) Schematic visualization of 9 HOG Orientation Bins (0°–180°, unsigned; arrows = gradient directions). Note that a gradient at 90° (arrow pointing straight up) comes from a horizontal edge, while a gradient at 0°/180° (arrow along the x-axis) comes from a vertical edge; (d) plot of global HOG orientation histogram (9 orientation bins, 0°–180°) in which each bar aggregates the total gradient magnitude for orientations falling into its 20° bin.}
\label{fig_2}
\end{figure}

HOG is a widely used feature descriptor for object detection and shape-based analysis, originally proposed by Dalal and Triggs for pedestrian detection \cite{dalal2005}. The core idea is to characterize local object appearance and shape by the distribution of edge orientations (gradients) (Fig.~\ref{fig_2}).
Given a grayscale image $I(x,y)$, the first step is to compute the gradient at each pixel using finite difference filters:
\begin{equation}
G_x = I(x+1,y) - I(x-1,y), \quad G_y = I(x,y+1) - I(x,y-1).
\end{equation}
From these gradients, the magnitude $M(x,y)$ and orientation $\theta(x,y)$ are computed as:
\begin{equation}
M(x,y) = \sqrt{G_x^2 + G_y^2}, \quad \theta(x,y) = \arctan2(G_y, G_x).
\end{equation}

The image is divided into small connected regions called cells (e.g., $8\times8$ pixels), and a histogram of gradient directions is computed within each cell, typically with 9 orientation bins spanning $0^\circ$ to $180^\circ$ (unsigned) (Fig.~\ref{fig_2}). Each pixel contributes to the histogram proportionally to its gradient magnitude.

To enhance robustness to illumination and contrast variations, the histograms are normalized across larger regions called blocks (e.g., $2\times2$ cells), using $L_2$ or $L_1$-norm normalization:
\begin{equation}
v \leftarrow \frac{v}{\sqrt{\|v\|_2^2 + \epsilon^2}},
\end{equation}
where $v$ is the concatenated vector of histograms in a block and $\epsilon$ is a small constant to avoid division by zero.

HOG features are translation-invariant at the cell level and capture edge and contour information that is highly discriminative for structured objects.
In addition, HOG remains a widely used handcrafted descriptor in applied vision pipelines due to its robustness in capturing gradient/edge structure. For example, Sun et al. used a vision plus SVM pipeline for train classification by converting weigh-in-motion (WIM) time series into images and extracting visual features for SVM-based categorization ~\cite{HOG1}. In agricultural machine vision, Xu et al. proposed a two-stage strawberry detection method that combines HSV-based candidate localization with an HOG+SVM classifier to handle slightly overlapping fruit ~\cite{HOG2}. The studies as above motivate HOG as a strong, task-relevant shape/edge baseline, which we include as one component in our hybrid feature stack alongside permutation-entropy complexity cues and LBP texture descriptors.

\subsection{Local Binary Pattern (LBP)}
LBP is a texture descriptor that encodes local spatial patterns by thresholding the neighborhood of each pixel. It was first introduced by Ojala et al. \cite{ojala2002} as a powerful and simple operator for texture classification.

For a pixel at position $(x,y)$, and a circular neighborhood of $P$ pixels at radius $R$, the LBP code is computed as:

\begin{equation}
\text{LBP}_{P,R}(x,y) = \sum_{p=0}^{P-1} s(g_p - g_c) \cdot 2^p,
\end{equation}
Here, $g_c$ denotes the gray value of the center pixel and $g_p$ denotes the gray value of the $p$-th neighbor (interpolated if it does not lie exactly on the pixel grid). The function
\[
s(z)=
\begin{cases}
1, & \text{if } z \ge 0,\\
0, & \text{otherwise,}
\end{cases}
\]
takes the value 1 when $z\ge0$ and 0 otherwise.

This results in a binary number (e.g., 8-bit for $P=8$) representing the local texture pattern. A histogram of LBP codes is then computed over local image regions and concatenated to form the final descriptor.

\begin{figure}[htbp!]
\centering
\includegraphics[width= 3.8 in]{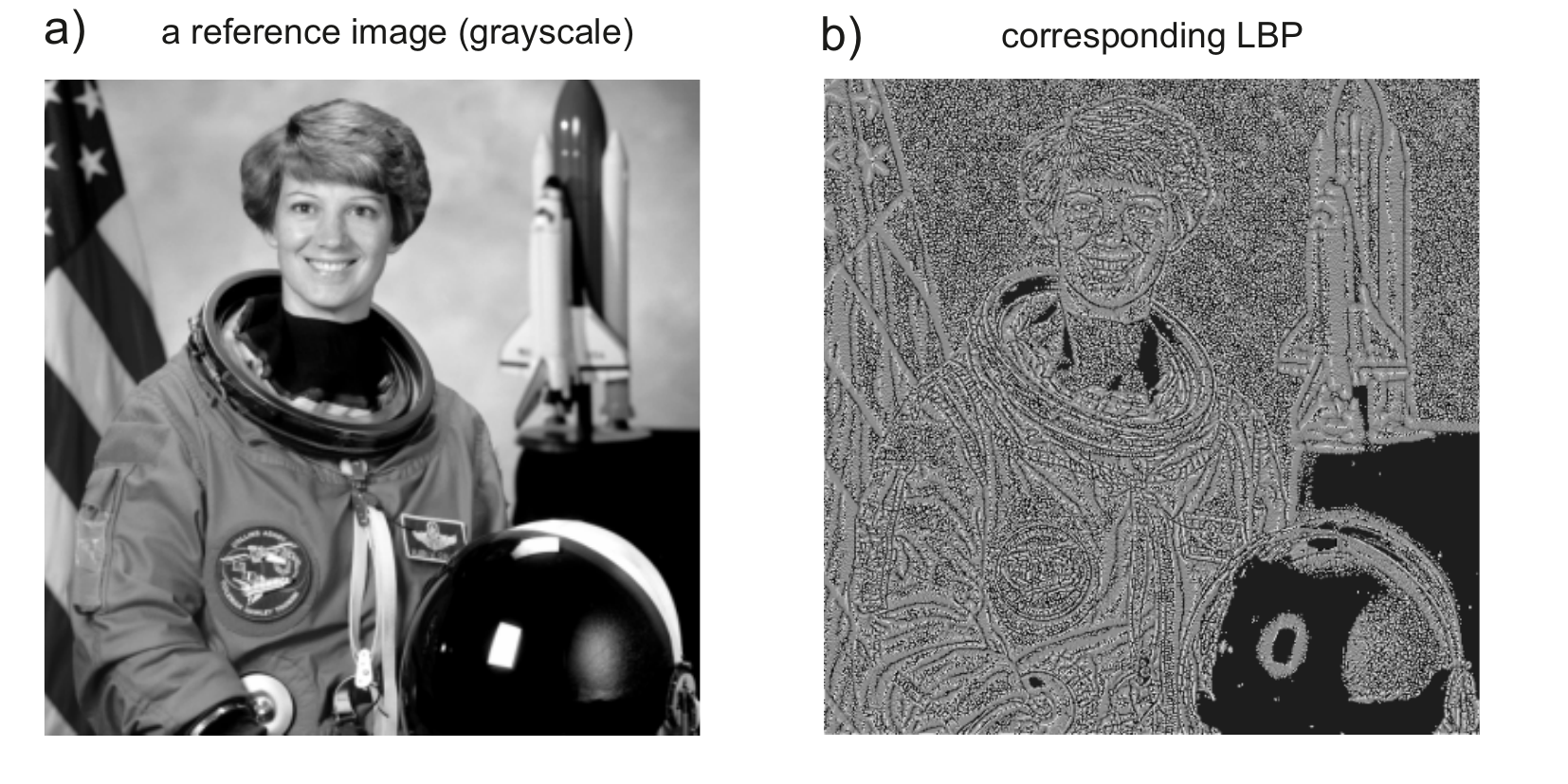}
\caption{Visualization of LBP from an image: (a) reference image (source: scikit-image); (b) corresponding LBP which is good at encoding fine-scaled local texture.}
\label{fig_3}
\end{figure}

LBP is highly discriminative for micro-patterns such as edges, corners, and spots, and it is invariant to monotonic gray-scale transformations, making it robust to illumination changes (Fig.~\ref{fig_3}).

\textbf{Combined Use in Feature Fusion:}
When used together, HOG and LBP provide complementary information: HOG captures global structure and shape, while LBP encodes fine-scaled local texture. Note that although HOG is more of an edge/shape descriptor, but its dense‐grid histogram approach can incidentally pick up some texture cues (Fig.~\ref{fig_2}). In contrasts, LBP is explicitly a texture descriptor (Fig.~\ref{fig_3}). Feature fusion of HOG and LBP has been shown to improve classification performance in various tasks, including face recognition, scene classification, and biomedical imaging \cite{sharifnejad2021fer}.

\section{The Feature Extraction Pipeline: The Fusion of Complexity, Correlation, Shape and Texture}\label{sec:FE_extraction}

\subsection{Feature Extraction Process and Dimensions}

\begin{figure*}[htbp!]
\centering
\includegraphics[width=7in]{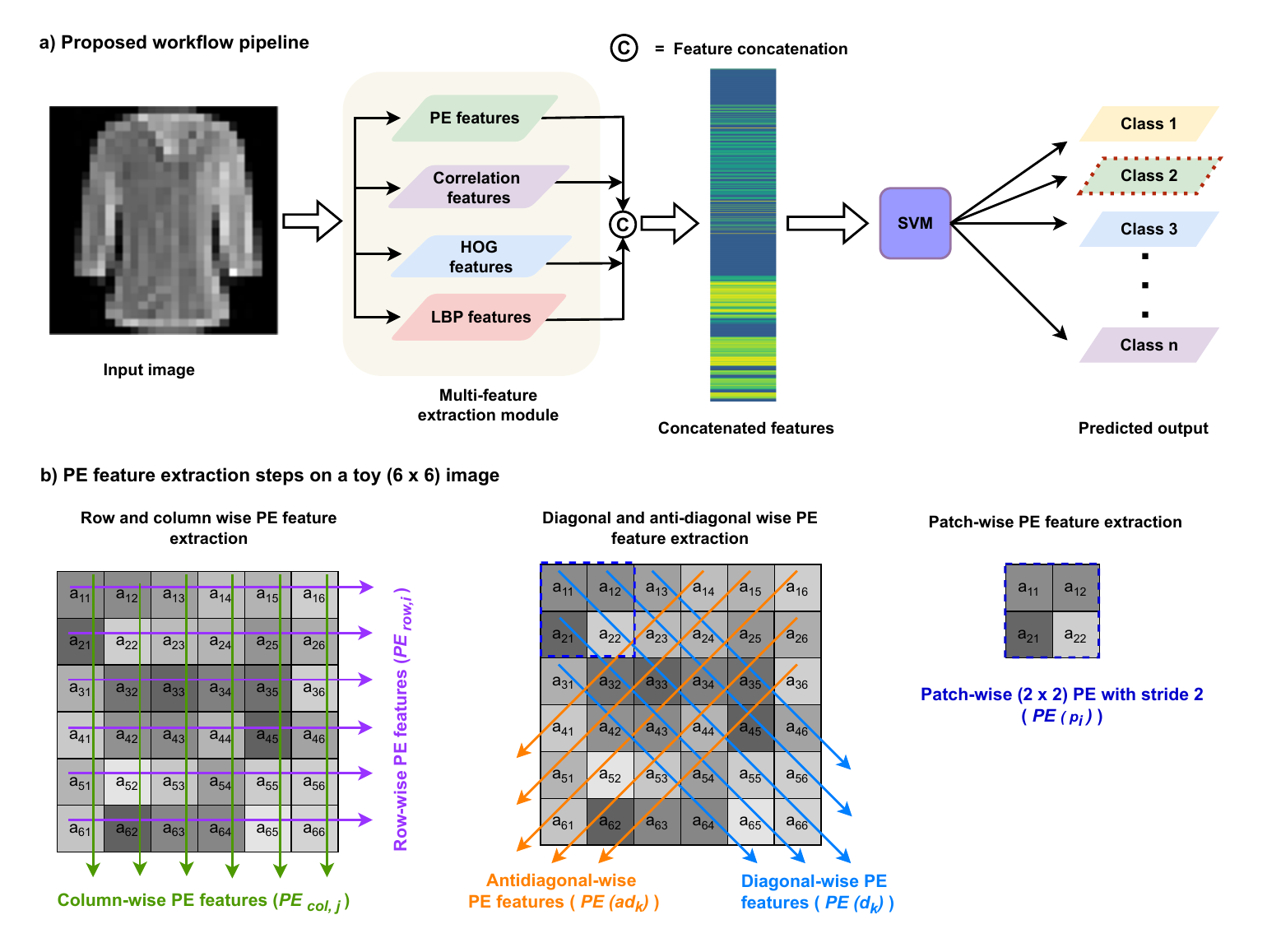}
\caption{ (a) Schematic overview of the proposed workflow pipeline. For each input image we extract permutation-entropy (PE) based features—computed along rows, columns, the diagonal and anti-diagonals, and on overlapping local patches. We combine them with row-/column-correlation based features, histogram of oriented gradients (HOG) and local binary pattern (LBP) image descriptors. The four feature sets are concatenated into a single feature vector and classified with a support vector machine (SVM). (b) Detailed illustration of the PE-based feature-extraction steps on a (6 × 6) toy image. PE is first evaluated for every row and column, followed by the diagonal (top-left → bottom-right) and anti-diagonal (top-right → bottom-left) directions. Note that some off-centre diagonals and anti-diagonals (towards corner) PEs are not computed because their reduced length may yield unreliable PE features or statistics. Finally, we compute patch-wise PE with (2 × 2) patches and with a stride length of 2 pixels. See Section \ref{sec:FE_extraction} for a complete description of the feature-extraction pipeline. }
\label{fig_4}
\end{figure*}

In this work, each image from standard benchmark datasets (such as Fashion-MNIST, KMNIST, EMNIST, and CIFAR-10) is transformed into a handcrafted feature vector composed of entropy-based, correlation-based, shape and texture-based image descriptors. Although dimensionality may vary slightly between datasets due to image size or content, the core extraction mechanisms remain the same (Fig.~\ref{fig_4}). In the following, we provide a general overview of the feature types and the methods used for their extraction.

\textbf{Row-wise PE}:  
Each row in the image is treated as a 1D signal. For each row vector $\mathbf{r}$, both the forward $PE(\mathbf{r})$ and reversed permutation entropy $PE(\mathbf{r}^{\text{rev}})$ are calculated, which means that each row of the image is treated as a one-dimensional signal: for a given row vector $\mathbf{r}$, the sequence of pixel values is interpreted first in the forward (left-to-right) direction and then in the reversed (right-to-left) direction, and the permutation entropy is computed for both resulting time series. Finally, the final geometric mean PE is calculated as follows:
    \[
        PE_{\text{row}}^{GM} = \sqrt{PE(\mathbf{r}) \cdot PE(\mathbf{r}^{\text{rev}})}
    \]
    yielding one feature per row. We do so for all possible rows.

 \textbf{Column-wise PE}: Analogously to the row-wise case, for each column vector $\mathbf{c}$ (note that this vector $\mathbf{c}$ can be treated as 1D signal/time series data) we compute the forward and reversed permutation entropies and combine them using geometric mean as follows:
    \[
        PE_{\text{col}}^{GM} = \sqrt{PE(\mathbf{c}) \cdot PE(\mathbf{c}^{\text{rev}})}
    \]
    yielding one feature per column.

\textbf{Diagonal PE}:  
We simplify the diagonal feature extraction as follows. Starting with the main diagonal (top-left to bottom-right), we also take the $K$ diagonals above it and the $K$ diagonals below it—so in total $2K + 1$ diagonals (Fig. 4b). For each diagonal line, we read its pixel values in order to form a one-dimensional sequence, then compute permutation entropy on that sequence and again on its reversed sequence. By taking the geometric mean of these two entropy values, we collapse each diagonal into a single feature . This reflects how complex the pixel arrangement is along that slanted direction.

    Thus, for each diagonal $d_k$ we have:
    \[
        PE_{\text{diag}}(k) = \sqrt{PE(d_k) \cdot PE(d_k^{\text{rev}})}
    \]

\textbf{Anti-Diagonal PE}: 
We handle anti-diagonals in exactly the same way. An anti-diagonal is simply a line of pixels running from the top-right corner down to the bottom-left corner (and any lines parallel to it; Fig. 4b). Concretely, we take the main anti-diagonal (top-right to bottom-left) and its $K$ offsets on either side, thus again yielding $2K+1$ lines—read each line’s pixel values into a one-dimensional sequence, compute permutation entropy on that sequence and on its reverse, then combine them via the geometric mean. This produces one scalar feature per anti-diagonal that, like the diagonal features, captures the complexity of pixel arrangements along that slanted orientation.

    Thus, for each anti-diagonal $ad_k$ we have:
    \[
        PE_{\text{anti-diag}}(k) = \sqrt{PE(ad_k) \cdot PE(ad_k^{\text{rev}})}
    \]
    
\textbf{Patch-wise PE}:  
The image is divided into overlapping patches (e.g., $4 \times 4$ with stride 2; Fig. 4b), and the PE is calculated for each patch:
    \[
        PE_{\text{patch}}(i) = PE(p_i)
    \]

\textbf{Row and Column Correlation}:  
Pearson correlation coefficients are computed between each pair of adjacent rows and adjacent columns:
    \[
        \rho_{i, i+1}^{\text{row}} = \text{corr}(\mathbf{r}_i, \mathbf{r}_{i+1}), \quad
        \rho_{j, j+1}^{\text{col}} = \text{corr}(\mathbf{c}_j, \mathbf{c}_{j+1})
    \]

\textbf{HOG features}:  
HOG features are computed using standard cell/block configurations (e.g., $4 \times 4$ pixels per cell) and a fixed number of orientation bins.
\\

\textbf{LBP features}:  
LBP descriptors are extracted with a fixed radius and number of sampling points (e.g., $R=2$, $P=16$), resulting in a histogram of uniform patterns.

\begin{table}
  \centering
  \caption{Toy \(6\times6\) image used to illustrate feature extraction. All pixel values $a_{ij}$ of the image are shown in the table}
  \label{images-1}
  \begin{tabular}{c|cccccc}
        & \multicolumn{1}{c}{\bf1} & \bf2    & \bf3    & \bf4    & \bf5    & \bf6    \\ \hline
    \bf1 & \(a_{11}\) & \(a_{12}\) & \(a_{13}\) & \(a_{14}\) & \(a_{15}\) & \(a_{16}\) \\
    \bf2 & \(a_{21}\) & \(a_{22}\) & \(a_{23}\) & \(a_{24}\) & \(a_{25}\) & \(a_{26}\) \\
    \bf3 & \(a_{31}\) & \(a_{32}\) & \(a_{33}\) & \(a_{34}\) & \(a_{35}\) & \(a_{36}\) \\
    \bf4 & \(a_{41}\) & \(a_{42}\) & \(a_{43}\) & \(a_{44}\) & \(a_{45}\) & \(a_{46}\) \\
    \bf5 & \(a_{51}\) & \(a_{52}\) & \(a_{53}\) & \(a_{54}\) & \(a_{55}\) & \(a_{56}\) \\
    \bf6 & \(a_{61}\) & \(a_{62}\) & \(a_{63}\) & \(a_{64}\) & \(a_{65}\) & \(a_{66}\) \\
  \end{tabular}
  
\end{table}

\noindent
\\
To illustrate the feature extraction process, we use a $6\times6$ toy image whose pixel values $a_{ij}$ are listed in Table~\ref{images-1}. We calculate the Row-wise PE, for each row $i$ ($i=1,\dots,6$), and form the sequence
$$
r_i = \bigl(a_{i1},\,a_{i2},\,\dots,\,a_{i6}\bigr),
$$
and compute its permutation entropy in the forward direction,
$$
PE\bigl(r_i\bigr),
$$
and in the reversed direction,
$$
PE\bigl(r_i^{\mathrm{rev}}\bigr),
\quad
r_i^{\mathrm{rev}} = (a_{i6},\,a_{i5},\,\dots,\,a_{i1}).
$$
Finally, the geometric‐mean PE feature for row $i$ is
$$
PE_{\mathrm{row},\,i}^{GM}
= \sqrt{\,PE\bigl(r_i\bigr)\times PE\bigl(r_i^{\mathrm{rev}}\bigr)\,}.
$$
Applying this to $i=1,\ldots,6$ produces the six row‐wise features
$\{PE_{\mathrm{row},1}^{GM},\,PE_{\mathrm{row},2}^{GM},\,\dots,\,PE_{\mathrm{row},6}^{GM}\}$.

\noindent
\\
In case of Column-wise PE, for each column $j=1,\dots,6$, define
$$
c_j = \bigl(a_{1j},\,a_{2j},\,\dots,\,a_{6j}\bigr),
c_j^{\mathrm{rev}} = (a_{6j},\,a_{5j},\,\dots,\,a_{1j}).
$$
Compute $PE(c_j)$ and $PE(c_j^{\mathrm{rev}})$, then form
$$
PE_{\mathrm{col},\,j}^{GM}
= \sqrt{\,PE(c_j)\,\times\,PE(c_j^{\mathrm{rev}})\,}\,, 
\quad j=1,\dots,6,
$$
yielding six column‐wise features $\{PE_{\mathrm{col},\,1}^{GM},\,PE_{\mathrm{col},\,2}^{GM},\,\dots,\,PE_{\mathrm{col},\,6}^{GM}\}$.

\noindent
\\
Let us now move to Diagonal-wise PE with $K=2$. For each offset $k=-2,-1,0,1,2$ define the diagonal

\begin{align*}
d_k &= \bigl(a_{p,\;p+k}\;\bigm|\;1\le p\le6,\;1\le p+k\le6\bigr), \\
d_k^{\mathrm{rev}} &= \bigl(a_{q,\;q+k}\;\bigm|\;q\text{ in reverse order}\bigr).
\end{align*}

{\small
\begin{align*}
d_{-2} &= (a_{31},\,a_{42}), 
    & d_{-2}^{\mathrm{rev}} &= (a_{42},\,a_{31}),\\
d_{-1} &= (a_{21},\,a_{32},\,a_{43},\,a_{54}), 
    & d_{-1}^{\mathrm{rev}} &= (a_{54},\,a_{43},\,a_{32},\,a_{21}),\\
d_{0}  &= (a_{11},\,a_{22},\,...,\,a_{66}),
    & d_{0}^{\mathrm{rev}}  &= (a_{66},\,a_{55},\,...,\,a_{11}),\\
d_{1}  &= (a_{12},\,a_{23},\,a_{34},\,a_{45},\,a_{56}),
    & d_{1}^{\mathrm{rev}}  &= (a_{56},\,a_{45},\,a_{34},\,a_{23},\,a_{12}),\\
d_{2}  &= (a_{13},\,a_{24},\,a_{35},\,a_{46}), 
    & d_{2}^{\mathrm{rev}}  &= (a_{46},\,a_{35},\,a_{24},\,a_{13}).
\end{align*}
}

\noindent
\\
Compute $PE(d_k)$ and $PE(d_k^{\mathrm{rev}})$ for five diagonal features.
$$
PE_{\mathrm{diag}}^{GM}(k)
= \sqrt{\,PE(d_k)\,\times\,PE(d_k^{\mathrm{rev}})\,}\,, 
\quad k=-2,\ldots,2,
$$

\noindent
\\
Similarly, for each $k=-2,-1,0,1,2$ define the anti‐diagonal
\begin{align*}
    ad_k = \bigl(a_{p,\;7-p+k}\;\bigm|\;1\le p\le6,\;1\le7-p+k\le6\bigr),
\\
ad_k^{\mathrm{rev}} = \bigl(a_{q,\;7-q+k}\;\bigm|\;q\text{ in reverse order}\bigr).
\end{align*}

\noindent
\\
Compute $PE(ad_k)$ and $PE(ad_k^{\mathrm{rev}})$, then for five anti‐diagonal features.
$$
PE_{\mathrm{anti}}^{GM}(k)
= \sqrt{\,PE(ad_k)\,\times\,PE(ad_k^{\mathrm{rev}})\,}\,, 
\quad k=-2,\ldots,2,
\\
\\
$$

\section{Support Vector Machine (SVM) for image classification }\label{sec:SVM}

\textbf{SVM} \cite{cortes1995support} are a robust class of supervised learning models that seek an optimal separating hyperplane by solving a regularized hinge loss minimization problem:
\begin{equation}
\min_{\mathbf{w},b} \frac{1}{2}\|\mathbf{w}\|^2 + C\sum_{i=1}^n \max(0, 1 - y_i(\mathbf{w}^T \phi(\mathbf{x}_i) + b)),
\end{equation}

\noindent
where, \(\mathbf{w}\in\mathbb{R}^d\) — The weight vector, normal to the separating hyperplane in feature space.

\noindent
\(b\in\mathbb{R}\) — The bias (intercept) term, which shifts the hyperplane away from the origin.

\noindent
\(\|\mathbf{w}\|^2\) — The squared Euclidean norm of \(\mathbf{w}\), penalizing large weights to encourage a wide margin.

\noindent
\(C>0\) — The regularization parameter that balances margin maximization (small \(\|\mathbf{w}\|\)) against misclassification errors; larger \(C\) increases the penalty on hinge‐loss violations.

\noindent
\(\phi:\mathbb{R}^p\to\mathcal{H}\) — An (often implicit) feature map into a (possibly high‐ or infinite‐dimensional) Hilbert space \(\mathcal{H}\). Kernels allow evaluation of inner products in \(\mathcal{H}\) without explicit computation of \(\phi\).

\noindent
\(y_i\in\{+1,-1\}\) — The true class label of example \(i\).

\noindent
\(\mathbf{x}_i\in\mathbb{R}^p\) — The input feature vector of example \(i\).

\noindent
\(\max\bigl(0,\,1 - y_i(\mathbf{w}^T\phi(\mathbf{x}_i)+b)\bigr)\) — The hinge loss for example \(i\): zero if correctly classified with margin at least 1, and growing linearly when the functional margin \(y_i(\mathbf{w}^T\phi(\mathbf{x}_i)+b)\) falls below 1.

In this work, we employ SVM not in their traditional raw pixel input setting \cite{ranzato2019robustness}, but in conjunction with a novel handcrafted feature extraction approach based on permutation entropy. To further enhance discriminative power, these entropy-based descriptors are combined with classical texture features: histogram of oriented gradients and local binary patterns. The result is a compact 780-dimensional feature vector that is interpretable, low-dimensional, and well-suited for classical machine learning models such as SVM.

Using this feature representation, we train SVM on four benchmark datasets: Fashion-MNIST, KMNIST, EMNIST (Balanced), and CIFAR-10. The use of handcrafted features allows SVM to operate effectively in a reduced-dimensional space, avoiding the scalability bottlenecks associated with high-dimensional kernel computations.

\section{ Benchmark Datasets: Fashion-MNIST, KMNIST, EMNIST, and CIFAR -10 }\label{sec:Dataset}

\subsection{Fashion-MNIST Dataset}
\textbf{Fashion-MNIST} is a benchmark dataset consisting of 70,000 grayscale images, each with a resolution of $28\times28$ pixels, categorized into 10 classes of clothing items: T-shirts/tops, trousers, pullovers, dresses, coats, sandals, shirts, sneakers, bags, and ankle boots \cite{xiao2017fashion}. The dataset is evenly split into 60,000 training images and 10,000 test images, with each class uniformly represented.

Despite its relatively small image size and lack of color information, Fashion-MNIST presents a meaningful challenge due to high intra-class variability and subtle inter-class differences. For example, distinguishing between shirts and T-shirts or between sneakers and ankle boots often requires sensitivity to texture, contour, and fine-grained shape differences. The images also exhibit varied orientations, partial occlusions, and diverse styles across classes.

These attributes make Fashion-MNIST a valuable resource for evaluating feature extraction methods that aim to capture structural and spatial patterns. Its grayscale format ensures that classification performance is driven primarily by shape and texture, rather than color, which aligns well with entropy-based approaches such as permutation entropy \cite{bandt2002permutation}.

\subsection{KMNIST Dataset}
\textbf{KMNIST} (Kuzushiji-MNIST) is a drop-in replacement for MNIST, created to encourage research into Japanese character recognition \cite{clanuwat2018deep}. It contains 70,000 $28\times28$ grayscale images of cursive Japanese (Kuzushiji) characters from 10 classes. Unlike Latin characters in MNIST or fashion items in Fashion-MNIST, the cursive nature of Kuzushiji introduces unique challenges such as high intra-class variability, stroke overlap, and complex calligraphic patterns.

KMNIST's higher visual complexity, despite its small size and grayscale format, presents an ideal middle ground between simplistic datasets like MNIST and natural image datasets like CIFAR-10. Its emphasis on subtle spatial structures makes it particularly compatible with entropy-based feature extraction approaches, which aim to quantify local complexity and variability within pixel arrangements.

\subsection{EMNIST Dataset}
\textbf{EMNIST} (Extended MNIST) expands upon the original MNIST dataset by including handwritten letters in addition to digits \cite{cohen2017emnist}. It consists of several splits; the most commonly used are EMNIST-Balanced and EMNIST-ByClass. The full dataset includes up to 814,255 $28\times28$ grayscale images of 62 classes (digits 0–9, uppercase A–Z, and lowercase a–z). EMNIST introduces more structural diversity and class imbalance, especially in the cases of handwritten letters where individual writing styles cause considerable intra-class variation.

From a feature extraction standpoint, EMNIST provides a broad and challenging landscape. The similar morphology of certain characters (e.g., `O' vs. `0', `I' vs. `l') creates nontrivial inter-class similarity, making it a valuable testbed for evaluating fine-grained structural descriptors such as permutation entropy.

\subsection{CIFAR-10 Dataset}
\textbf{CIFAR-10} is a benchmark dataset for general-purpose object classification in natural scenes \cite{krizhevsky2009learning}. It comprises 60,000 $32\times32$ color images across 10 object categories: airplane, automobile, bird, cat, deer, dog, frog, horse, ship, and truck. Unlike the previously discussed grayscale datasets, CIFAR-10 images are in full color and exhibit substantial variability in lighting, viewpoint, background, and object deformation.

The presence of color channels increases dimensionality and realism, but also introduces noise that can obscure structural patterns. While deep convolutional networks can achieve over 95\% accuracy on CIFAR-10 \cite{he2016deep}, such performance is often driven by color and texture cues rather than purely structural features. For this reason, CIFAR-10 serves as a more difficult and less isolated setting for entropy-based feature extraction, requiring careful preprocessing or channel-wise analysis to focus on spatial complexity.

\textbf{Comparative Analysis of the Dataset:}
The selected datasets—Fashion-MNIST, KMNIST, EMNIST, and CIFAR-10—represent a diverse spectrum of visual complexity. Fashion-MNIST and KMNIST retain the simplicity of small grayscale images while introducing semantic and structural diversity. EMNIST increases the class space and handwriting variability, and CIFAR-10 brings in full-color imagery with real-world variability.

For our investigation of permutation entropy-based feature extraction, Fashion-MNIST, KMNIST, and EMNIST offer the most controlled yet challenging environments. Their grayscale format isolates structural and textural elements, making them ideal for analyzing spatial complexity. In contrast, CIFAR-10 introduces naturalistic noise and color-based variation, highlighting the limitations and potential adaptations needed when applying entropy-based techniques in more complex visual domains.

\section{Image Classification Results and Discussion}\label{sec:Results}
\subsection{Results on Fashion-MNIST Dataset}

\begin{figure}[htbp!]
\centering
\includegraphics[width= 3.8 in]{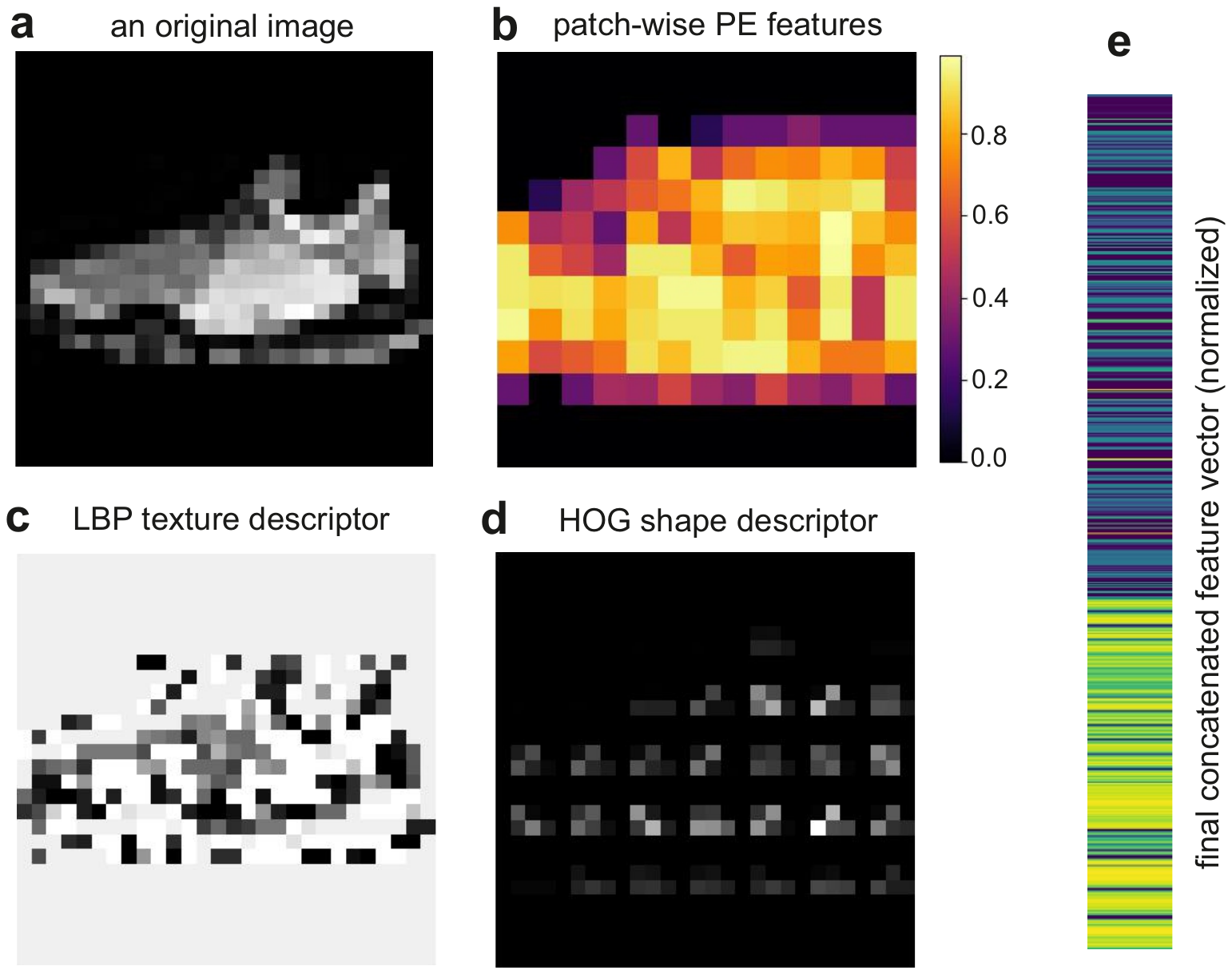}
\caption{Visualization of extracted features from a single Fashion-MNIST sample: (a) original 28×28 grayscale image; (b) patch-wise permutation-entropy (PE) heatmap computed over 4×4 windows with stride 2; (c) local binary pattern (LBP) texture map using P=16, R=2; (d) Histogram of Oriented Gradients (HOG) visualization with 4×4 pixels per cell; and (e) finally a normalized, concatenated feature vector used as input to the SVM classifier.}
\label{fig_5}
\end{figure}

We apply the proposed feature extraction pipeline (Fig.~\ref{fig_4}) to the \textbf{Fashion-MNIST} dataset, that comprises $28 \times 28$ grayscale images of fashion articles. Based on our method, each image is converted into a 780-dimensional feature vector, comprising entropy-based, correlation-based, shape-based and texture-based features (Fig.~\ref{fig_5}) as detailed below:

\textbf{Row-wise PE:} 28 rows → 28 features;
\textbf{Column-wise PE:} 28 columns → 28 features;
\textbf{Row Correlation:} 27 adjacent row pairs → 27 features;
\textbf{Column Correlation:} 27 adjacent column pairs → 27 features;
\textbf{Diagonal PE:} 21 diagonals ($k = -10$ to $10$) → 21 features;
\textbf{Anti-diagonal PE:} 21 diagonals on horizontally flipped image → 21 features;
\textbf{Patch-wise PE:} 169 overlapping patches of size $4 \times 4$ with stride 2 → 169 features;
\textbf{HOG:} 441 features using $4 \times 4$ pixel cells and 9 orientation bins;
\textbf{LBP:} 18-bin histogram with $P=16$, $R=2$ (uniform pattern) → 18 features.

\begin{table}[htbp!]
	\centering
    \caption{Breakdown of handcrafted feature types used in Fashion-MNIST.}
	\label{tab:feature_breakdown}
	\begin{tabularx}{\linewidth}{@{}lXr@{}}
		\toprule
		\textbf{Feature Type} & \textbf{(Count rationale)} & \textbf{Features} \\
		\midrule
		Row-wise PE            & (28 rows) & 28 \\
		Column-wise PE         & (28 columns) & 28 \\
		Row Correlation        & (27 adjacent row pairs) & 27 \\
		Column Correlation     & (27 adjacent column pairs) & 27 \\
		Diagonal PE            & ($k=-10,\ldots,10$) & 21 \\
		Anti-diagonal PE       & (flip + $k=-10,\ldots,10$) & 21 \\
		Patch-wise PE          & ($4{\times}4$, stride 2 $\Rightarrow 13{\times}13$) & 169 \\
		HOG                    & ($4{\times}4$ cells, 9 bins $\Rightarrow 7{\times}7{\times}9$) & 441 \\
		LBP                    & ($P=16$, $R=2$ $\Rightarrow P{+}2$ bins) & 18 \\
		\midrule
		\textbf{Total}         &  & \textbf{780} \\
		\bottomrule
	\end{tabularx}
\end{table}

This 780-dimensional feature vector (Table~\ref{tab:feature_breakdown}) is standardized and used as input to the classification models. Similar extraction procedures are applied to other datasets (KMNIST, EMNIST, CIFAR-10), with minor adjustments to patch count or image resolution as necessary. The resulting feature matrices had shapes $(60000, 780)$ for training and $(10000, 780)$ for testing. An SVM classifier with RBF kernel was tuned using 3-fold cross-validation over a grid of 12 $(C, \gamma)$ combinations, totaling 36 fits. The best hyperparameters obtained were $C=10$ and $\gamma=0.001$.The best cross-validation accuracy achieved during tuning was \textbf{90.44\%}, and the final test accuracy using the best SVM model was \textbf{91.23\%}. Table~\ref{tab:svm_comparison} presents a comparison of our results with previously reported SVM-based approaches on the Fashion-MNIST dataset that did not employ deep learning or handcrafted descriptors of comparable complexity.


\begin{table}[htbp]
	\centering
    	\caption{Comparison with existing Classical ML-based methods on Fashion-MNIST (classical ML only)}
	\label{tab:svm_comparison}
	\begin{tabularx}{\linewidth}{@{}lXr@{}}
		\toprule
		\textbf{Method} & \textbf{Feature Type} & \textbf{Accuracy} \\
		\midrule
		\midrule
		Logistic regression \cite{xiao2017fashion} 
		  & Raw $28 \times 28$ flattened 
		  & 84.2\% \\
		\addlinespace
		Random forest \cite{fashionmnistweb} 
		  & Raw pixels 
		  & 87.3\% \\
		\addlinespace
		KNN Classifier\cite{xiao2017fashion} 
		  & Raw pixels
		  & 85.4\% \\
		\addlinespace
        Linear SVM \cite{xiao2017fashion} 
		  & Raw $28 \times 28$ flattened 
		  & 83.6\% \\
		\addlinespace
        RBF SVM \cite{fashionmnistweb} 
		  & Raw $28 \times 28$ flattened 
		  & 89.7\% \\
		\addlinespace
		\textbf{PE/Corr + HOG + LBP + SVM} 
		  & 780 handcrafted features 
		  & \textbf{91.23\%} \\
		\bottomrule
	\end{tabularx}
\end{table}

\begin{figure*}[!t]
\centering
\includegraphics[width=7in]{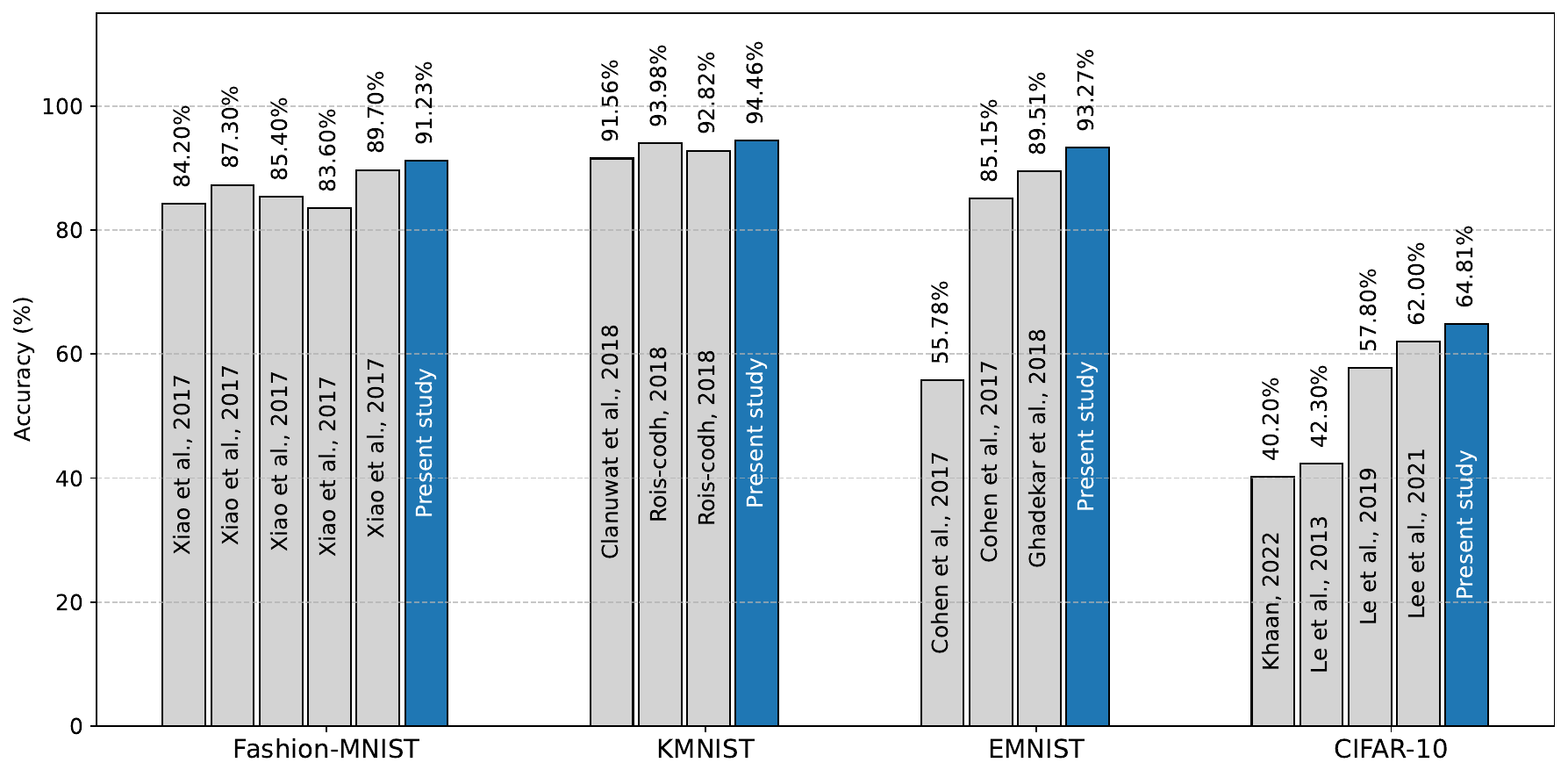}
\caption{Test accuracy comparison across benchmark datasets between previous classical machine learning methods and our approach (present study) that combines handcrafted feature extraction with SVM. Our method consistently outperforms traditional classical ML-based methods and feature descriptors, demonstrating its effectiveness and efficiency across Fashion-MNIST, KMNIST, EMNIST, and CIFAR-10 dataset.}
\label{fig_6}
\end{figure*}

The proposed method significantly improves over conventional SVM models trained on raw pixel intensities or texture-only features. By integrating entropy-based and correlation-based descriptors with traditional texture features like HOG and LBP, the model captures both statistical randomness and structural dependencies within the image. This comprehensive feature set enables better generalization and class separability, especially for visually similar classes in Fashion-MNIST. Compared to the best reported classical SVM accuracy of 89.7\%~\cite{fashionmnistweb}, our method achieves a relative improvement of approximately \textbf{1.6\%} in absolute accuracy. This is a notable gain considering no deep learning models were used. Moreover, the method remains computationally feasible and inexpensive, with model training times ranging between 8–67 minutes depending on hyperparameters, which is acceptable for non-deep learning pipelines. This validates the effectiveness of the proposed handcrafted feature extraction method, and achieves superior performance compared to traditional SVM approaches reported in literature (Table~\ref{tab:svm_comparison} and Fig.~\ref{fig_6}).

\begin{table}[t]
	\centering
	\caption{Fashion-MNIST ablation (60k/10k split). Accuracy is reported using a fixed RBF-SVM configuration ($C{=}10$, $\gamma{=}0.001$) for a controlled comparison across feature sets.}
	\label{tab:ablation_fashion_full_short}
	\begin{tabularx}{\linewidth}{@{}Xrr@{}}
		\toprule
		\textbf{Feature set} & \textbf{Dim.} & \textbf{Acc. (\%)} \\
		\midrule
		\midrule
		PE/Corr 
		  & 321 
		  & 87.31 \\
		\addlinespace
		HOG 
		  & 441 
		  & 89.28 \\
		\addlinespace
		LBP 
		  & 18 
		  & 55.34 \\
		\addlinespace
		HOG+LBP 
		  & 459 
		  & 89.90 \\
		\addlinespace
		PE/Corr+HOG 
		  & 762 
		  & 90.87 \\
		\addlinespace
		PE/Corr+LBP 
		  & 339 
		  & 87.74 \\
		\addlinespace
		\textbf{Full (PE/Corr+HOG+LBP)} 
		  & \textbf{780} 
		  & \textbf{91.23} \\
		\bottomrule
	\end{tabularx}
\end{table}

\begin{table}[t]
	\centering
	\caption{Sensitivity of the PE/Corr feature block (321-D) to the permutation entropy parameters: embedding dimension $m$ and delay $\tau$ on Fashion-MNIST (60k/10k split). Accuracy is reported using a fixed RBF-SVM configuration ($C{=}10$, $\gamma{=}0.001$) for a controlled comparison. \textit{Note: $m$ is the PE embedding dimension (pattern length); the “2”  indicates that the header cell spans two rows for formatting and is not $2{\times}m$.}}
	\label{tab:pe_sensitivity_fashion}
	\begin{tabularx}{\linewidth}{@{}c@{\hspace{10mm}}*{3}{>{\centering\arraybackslash}X}@{}}
		\toprule
		\multirow{2}{*}{\textbf{$m$}} & \multicolumn{3}{c}{\textbf{$\tau$}} \\
		\cmidrule(l){2-4}
		 & \textbf{1} & \textbf{2} & \textbf{3} \\
		\midrule
		\midrule
		3 & 86.88 & 87.29 & 87.06 \\
		\addlinespace
		4 & 87.31 & 86.91 & 86.92 \\
		\addlinespace
		5 & 86.96 & 86.69 & 86.80 \\
		\bottomrule
	\end{tabularx}
\end{table}

Tables~\ref{tab:ablation_fashion_full_short} and~\ref{tab:pe_sensitivity_fashion} provide parameter sensitivity for the proposed design choices on Fashion-MNIST. The ablation in Table~\ref{tab:ablation_fashion_full_short} shows that the PE/Corr block alone is competitive (87.31\%), while HOG contributes the strongest individual gain (89.28\%); combining PE/Corr with HOG improves accuracy to 90.87\%, and the full descriptor achieves the best performance (91.23\%), indicating that ordinal-entropy/correlation statistics provide information complementary to gradient-based texture cues. To address sensitivity of permutation entropy to its core parameters,
Table~\ref{tab:pe_sensitivity_fashion} varies the embedding dimension $m$ and delay $\tau$. Performance is stable across the tested range ($m\in{3,4,5}$ and $\tau\in{1,2,3}$), with a small spread of approximately 0.6 percentage points, suggesting that the PE/Corr representation is not overly sensitive to moderate changes in $(m,\tau)$ and supporting the robustness of the chosen default configuration.

For comparison of performance with previous tensor-based HOG studies ~\cite{vo2015tensorhog} ~\cite{zhu2023nonlinearstmhog}, we additionally evaluated HOG as a tensor of size $7 \times 7 \times 9$ (blocks$_\text{row}$ $\times$ blocks$_\text{col}$ $\times$ orientations) within the same hybrid feature stack (Entropy + HOG + LBP). As shown in Table~\ref{tab:tensorhog_comparison}, naive tensor flattening yields essentially the same performance as standard vector-HOG (91.19\% vs. 91.23\%), consistent with both representations containing the same HOG coefficients. MPCA (ranks $(7,7,9)$) slightly reduces accuracy (90.57\%), and a tensor-aware STM baseline (OVR, CP rank 8) achieves 72.94\% accuracy; thus, these tensor-based baselines do not improve over the tuned RBF-SVM reference in our setting.

\begin{table}[htbp]
	\centering
	\caption{Tensor-based HOG formulations under same hybrid feature stack on Fashion-MNIST.}
	\label{tab:tensorhog_comparison}
	\begin{tabularx}{\linewidth}{@{}lXr@{}}
		\toprule
		\textbf{Method} & \textbf{HOG Representation / Tensor Method} & \textbf{Accuracy} \\
		\midrule
		\midrule
		Baseline (Vector-HOG + SVM)
		  & Vector-HOG within 780 handcrafted features
		  & \textbf{91.23\%} \\
		\addlinespace
		Tensor-HOG (flatten) + SVM
		  & Tensor-HOG ($7 \times 7 \times 9$) flattened (same coefficients)
		  & 91.09\% \\
		\addlinespace
		Tensor-HOG + MPCA + SVM
		  & MPCA ranks $(7,7,9)$ (core dim = 441)
		  & 90.57\% \\
		\addlinespace
		Tensor-aware classifier (STM-OVR)
		  & STM-OVR (CP rank = 8) on tensor-HOG; Entropy+LBP unchanged
		  & 72.94\% \\
		\bottomrule
	\end{tabularx}
\end{table}

\subsection{Results on KMNIST Dataset}

We apply the same feature extraction pipeline (Fig.~\ref{fig_4}) to the KMNIST dataset, which consists of grayscale images of Japanese Kuzushiji characters. Unlike Fashion-MNIST, KMNIST has more complex shapes and strokes, but our combination of permutation entropy (PE), correlation features, Histogram of Oriented Gradients (HOG), and Local Binary Patterns (LBP) successfully captured the intrinsic texture and structural information. The resulting feature matrices had shapes $(60000, 780)$ for training and $(10000, 780)$ for testing

We performed hyperparameter tuning of an SVM with an RBF kernel via grid search using 3-fold cross-validation. The best parameters were found to be $C=200$ and $\gamma=0.001$, yielding a cross-validation accuracy of \textbf{97.55\%}. On the held-out test set, the tuned RBF SVM achieved a test accuracy of \textbf{94.46\%}, outperforming previous classical machine learning SVM approaches on KMNIST that used raw pixels or simpler handcrafted features. Table~\ref{tab:kmnist_svm_comparison} summarizes the comparison.

The improvement can be attributed to the rich feature representation that combines multiple complementary descriptors. Permutation entropy captures the complexity and irregularity of pixel intensity patterns, correlation features encode spatial relationships, and HOG and LBP extract local shape and texture patterns. 


\begin{table}[htbp]
	\centering
    	\caption{Comparison with existing Classical ML-based methods on KMNIST (classical ML only)}
	\label{tab:kmnist_svm_comparison}
	\begin{tabularx}{\linewidth}{@{}lXr@{}}
		\toprule
		\textbf{Method} & \textbf{Feature Type} & \textbf{Accuracy} \\
		\midrule
		\midrule
		4-Nearest Neighbour Baseline \cite{clanuwat2018deep}
		  & Raw pixels
		  & 91.56\% \\
		\addlinespace
		PCA + 4-KNN \cite{rois2020svm_kmnist}
		  & Raw pixels
		  & 93.98\% \\
		\addlinespace
		Tuned SVM \cite{rois2020svm_kmnist}
		  & Raw pixels
		  & 92.82\% \\
		\addlinespace
		\textbf{PE/Corr + HOG + LBP + SVM}
		  & 780 handcrafted features
		  & \textbf{94.46\%} \\
		\bottomrule
	\end{tabularx}
\end{table}

\subsection{Results on EMNIST Dataset}

We also tested our handcrafted feature extraction pipeline  to the EMNIST Letters dataset, generating feature vectors of shape (124800, 780) for training and (20800, 780) for testing. Using a 3-fold cross-validation with grid search over a range of SVM hyperparameters, the best model was found with parameters $C=10$ and $\gamma=0.001$. The best cross-validation accuracy achieved was \textbf{92.83\%}, while the tuned SVM attained a test accuracy of \textbf{93.27\%}.

As shown in Table~\ref{tab:emnist_svm_comparison}, our method outperforms existing classical SVM pipelines applied to EMNIST datasets. The results demonstrate the effectiveness of combining multiple handcrafted features for the EMNIST handwritten character recognition task. The performance is consistent with or better than previously reported classical SVM-based methods on EMNIST, which typically utilize individual feature types such as raw pixels or HOG descriptors.

\vspace{2mm}

\begin{table}[htbp]
	\centering
    	\caption{Comparison with existing Classical ML-based methods on EMNIST (classical ML only)}
	\label{tab:emnist_svm_comparison}
	\begin{tabularx}{\linewidth}{@{}lXr@{}}
		\toprule
		\textbf{Method} & \textbf{Feature Type} & \textbf{Accuracy} \\
		\midrule
		\midrule
        Linear classifier \cite{cohen2017emnist}
		  & Raw pixels
		  & 55.78\% \\
		\addlinespace
		OPIUM\cite{cohen2017emnist}
		  & Raw pixels
		  & 85.15\% \\
		\addlinespace
		DWT-DCT + SVM \cite{ghadekar2018dwt_dct}
		  &  Extracted features
		  & 89.51\% \\
		\addlinespace

		\textbf{PE/Corr + HOG + LBP + SVM}
		  & 780 handcrafted features
		  & \textbf{93.27\%} \\
		\bottomrule
	\end{tabularx}
\end{table}

\subsection{Results on CIFAR-10 Coloured Dataset}

To apply our handcrafted feature pipeline to CIFAR-10, we extended our method to handle RGB color channels. Each image was split into its red, green, and blue channels. Features were extracted independently from each channel using  the same feature extraction pipeline (Fig.~\ref{fig_4}). Each channel contributed a set of entropy and texture features (total 427 features), which were concatenated to form the final feature vector (yielding 1281 = 3 × 427 features). In total, this resulted in a 1281-dimensional feature vector per image, as shown in Table~\ref{tab:cifar10_feature_breakdown}. We performed hyperparameter tuning of an SVM with an RBF kernel via grid search using 3-fold cross-validation. The best parameters were found to be $C=100$ and $\gamma=0.001$, yielding a cross-validation accuracy of 61.51\%. On the held-out test set, the tuned SVM achieved an accuracy of \textbf{64.81\%}.

\begin{table}[htbp!]
	\centering
    \caption{Breakdown of handcrafted feature types used in CIFAR-10 (per RGB channel and total).}
	\label{tab:cifar10_feature_breakdown}
	\begin{tabularx}{\linewidth}{@{}lXrr@{}}
		\toprule
		\textbf{Feature Type} & \textbf{(Count rationale)} & \textbf{Per-channel} & \textbf{Total (R+G+B)} \\
		\midrule
        \midrule
		Row-wise PE            & (32 rows) & 32  & 96  \\
		Column-wise PE         & (32 columns) & 32  & 96  \\
		Row Correlation        & (31 adjacent row pairs) & 31  & 93  \\
		Column Correlation     & (31 adjacent column pairs) & 31  & 93  \\
		Main-diagonal PE       & ($8{\times}8$, stride 4 $\Rightarrow 7{\times}7$) & 49  & 147 \\
		Anti-diagonal PE       & (flip + $8{\times}8$, stride 4 $\Rightarrow 7{\times}7$) & 49  & 147 \\
		Patch-wise PE          & ($8{\times}8$, stride 4 $\Rightarrow 7{\times}7$) & 49  & 147 \\
		HOG                    & ($8{\times}8$ cells, 9 bins $\Rightarrow 4{\times}4{\times}9$) & 144 & 432 \\
		LBP                    & ($P=8$, $R=1$ $\Rightarrow P{+}2$ bins) & 10  & 30  \\
		\midrule
		\textbf{Total}         &  & \textbf{427} & \textbf{1281} \\
		\bottomrule
	\end{tabularx}
\end{table}


\begin{table}[htbp]
	\centering
    	\caption{Comparison with existing Classical ML-based methods on CIFAR-10 (classical ML only)}
	\label{tab:cifar_svm_comparison}
	\begin{tabularx}{\linewidth}{@{}lXr@{}}
		\toprule
		\textbf{Method} & \textbf{Feature Type} & \textbf{Accuracy} \\
		\midrule
		\midrule
		SVM + PCA \cite{osamakhaan2022cifar_svm_pca}
		  & Top 100 PCA components
		  & 40.2\% \\
        Linear SVM  \cite{le2013fastfood}
		  & 32 x 32 raw pixels
		  & 42.3\% \\
		\addlinespace
        SVM + Logistic Regression \cite{le2019cifar_svm_ensemble}
		  & SVM on raw pixels
		  & 57.8\% \\
		\addlinespace
		HOG + PCA + RBF SVM \cite{lemuel2021cifar_hog_pca_svm}
		  & Grayscaling and HOG
		  & 62.0\% \\
		\addlinespace
		\textbf{PE/Corr + HOG + LBP + SVM}
		  & 1281 handcrafted RGB features
		  & \textbf{64.81\%} \\
		\bottomrule
	\end{tabularx}
\end{table}

As shown in Table~\ref{tab:cifar_svm_comparison}, our method outperforms existing classical SVM pipelines applied to CIFAR-10. By combining entropy-based complexity measures (including patch-based permutation entropy), correlation structure, edge and texture descriptors (HOG and LBP), we capture richer information than standard approaches that rely solely on edge or intensity features.

\noindent\textbf{Color space and channel ablation on CIFAR-10.}
Using our original RGB configuration (entropy+HOG+LBP on all RGB channels) we obtain \textbf{64.81\%} accuracy. Additionaly, we evaluated channel-selective and alternative color-space variants: restricting features to ``structural'' channels (YCbCr $Y$ and HSV $V$) yields \textbf{60.42\%}, and extracting entropy across all YCbCr channels (with HOG/LBP on $Y$) yields \textbf{61.62\%}. These results suggest that CIFAR-10 benefits from chroma information; discarding or limiting color channels can reduce discriminative power.

A different study by Coates et al.\ \cite{coates2011analysis} extracted a 4\,000-dimensional feature vector by learning single-layer convolutional filters via K-means and simple pooling, achieving 79.6\% accuracy on CIFAR-10. However, their approach incurs a significant offline training cost and yields features that are difficult to interpret. In contrast, our permutation-entropy + HOG + LBP pipeline produces a compact, lightweight feature descriptor that can be computed in a fully feed-forward manner without any filter learning, and whose individual components (entropy, gradient histograms, binary patterns) offer clear, intuitive interpretations. This interpretability and efficiency comes with a modest trade-off in accuracy, making our method particularly well-suited to applications with limited compute or a need for transparent feature representations.

\section{Future Research Direction and Preliminary Results}
\subsection{Neural Networks and Hand-crafted Features}
To study how the proposed handcrafted descriptors interact with modern convolutional features, we consider an additional experiment on CIFAR-10 using a lightweight CNN backbone and a hybrid (deep + classical ) representation. Further, we chose the CIFAR-10 dataset to study this idea because it is a standard, moderately challenging natural image benchmark that is widely used to evaluate lightweight CNNs and allows fair comparison with both classical and deep learning methods.

As an efficient deep baseline, we adopt MobileNetV2, a compact architecture based on depthwise separable convolutions with inverted residual blocks and linear bottlenecks. The network is initialized with ImageNet-pretrained weights and adapted to CIFAR-10 by replacing the final classification layer with a 10-class head. From the original CIFAR-10 training set (50{,}000 images), we construct a train/validation split using a tail split of 45{,}000 training and 5{,}000 validation images, and evaluate on the standard 10{,}000-image test set. CIFAR-10 images are resized to $224\times224$ pixels and normalized with ImageNet mean and variance to match the backbone’s default input conditions.

The model is fine-tuned using Adam with cross-entropy loss for 20 epochs, with a batch size of 64. We track validation accuracy and select the best checkpoint for final evaluation. This MobileNetV2 configuration achieves a test accuracy of \textbf{92.43\%}, with a total fine-tuning time of 23{,}054\,s. The penultimate layer produces 1280-dimensional embeddings, which we export for all train/validation/test samples.

\begin{table}[t]
	\centering
	\caption{MobileNetV2 transfer-learning configuration on CIFAR-10.}
	\label{tab:mobilenet_setup}
	\begin{tabularx}{\linewidth}{@{}lXr@{}}
		\toprule
		\textbf{Setting} & \textbf{Value} & \textbf{Notes} \\
		\midrule
		\midrule
		Backbone
		  & MobileNetV2
		  & ImageNet-pretrained \\
		\addlinespace
		Input size
		  & $224\times224$
		  & Resized CIFAR-10 \\
		\addlinespace
		Train / Val / Test
		  & 45k / 5k / 10k
		  & Tail split on train \\
		\addlinespace
		Optimizer
		  & Adam
		  & Cross-entropy loss \\
		\addlinespace
		Epochs / Batch size
		  & 20 / 64
		  & Fine-tuning \\
		\addlinespace
		Embedding dimension
		  & 1280
		  & Penultimate layer \\
		\addlinespace
		Train time
		  & 23{,}054\,s
		  & CPU/GPU setup as in Sec.~X \\
		\addlinespace
		Test accuracy
		  & \textbf{92.43\%}
		  & Best val checkpoint \\
		\bottomrule
	\end{tabularx}
\end{table}

On the same CIFAR-10 splits, we compute our handcrafted feature vector, resulting in a 1281-dimensional descriptor per image. For each sample, we concatenate the MobileNetV2 embedding (1280 dimensions) with the handcrafted descriptor (1281 dimensions), forming a fused feature vector of dimension 2561. This final feature vector is hybrid in nature because it combines learned deep embeddings from MobileNetV2 with explicitly defined handcrafted descriptors bringing together semantic representation and interpretable edge/texture/complexity cues in a single feature space.

Finally, a linear SVM with squared hinge loss is trained in this fused space. The hyperparameters (regularization parameter $C$, tolerance, and class weighting) are tuned by using the Optuna-based hyperparameter search in the training/validation split. The best configuration (class\_weight = balanced, $C \approx 6.7\times10^{-4}$, tol $\approx 4.9\times10^{-5}$) yields a test accuracy of \textbf{93.10\%}, with an SVM training time of 313\,s.

\begin{table}[t]
	\centering
	\caption{Deep baseline vs.\ hybrid deep + handcrafted fusion on CIFAR-10.}
	\label{tab:mobilenet_fusion_results}
	\begin{tabularx}{\linewidth}{@{}lXr@{}}
		\toprule
		\textbf{Setting} & \textbf{Value} & \textbf{Notes} \\
		\midrule
		\midrule

		\multicolumn{3}{@{}l@{}}{\textbf{MobileNetV2 (fine-tuned)}} \\
		\addlinespace
		Feature dimension
		  & 1{,}280
		  & Deep embedding \\
		\addlinespace
		Classifier
		  & Softmax head
		  & End-to-end training \\
		\addlinespace
		Train time
		  & 23{,}054\,s
		  & As measured in our setup \\
		\addlinespace
		Test accuracy
		  & 92.43\%
		  & CPU-only \\

		\addlinespace
		\addlinespace
		\multicolumn{3}{@{}l@{}}{\textbf{MobileNetV2 + handcrafted fusion}} \\
		\addlinespace
		Feature dimension
		  & 2{,}561
		  & 1{,}280 (deep) + 1{,}281 (handcrafted) \\
		\addlinespace
		Classifier
		  & SVM
		  & Trained on fused features \\
		\addlinespace
		Train time
		  & 313\,s
		  & Classifier training only \\
		\addlinespace
		Test accuracy
		  & \textbf{93.10\%}
		  & CPU-only \\

		\bottomrule
	\end{tabularx}
\end{table}

The hybrid representation improves CIFAR-10 performance from 92.43\% (MobileNetV2 fine-tuning alone) to 93.10\% (MobileNetV2 embeddings fused with handcrafted features and classified by SVM), corresponding to a +0.67 percentage-point gain. At this accuracy level, where reaching above 90\% on CIFAR-10 is already non-trivial, such an improvement is meaningful and suggests that the proposed handcrafted descriptors provide complementary information to the deep embeddings. As a modest but meaningful gain at this accuracy level, this result is encouraging, and future work may explore similar deep + handcrafted fusion on a few other datasets, although a full-scale exploration is beyond the scope of this paper.

\subsection{Understanding efficiency}
In this subsection, we provide a quantitative comparison of the computational cost of the proposed handcrafted-feature pipeline and a deep learning (CNN) baseline we saw previously. We report separate timing measurements for feature extraction and classifier training on CIFAR-10, and we analyze how these components contribute to the overall efficiency of purely classical, purely deep, and hybrid deep+handcrafted configurations.

Here, the experiments were conducted on a high-performance Linux workstation running Ubuntu 22.04.5 LTS with an Intel Xeon Platinum 8280 CPU (112 cores, 4.0 GHz) and 772,640 MiB of system memory, using CPU-only computation throughout.

Let us state the results. From a classical machine learning perspective, the SVM trained on our handcrafted features and achieving 64.81\% accuracy on CIFAR-10 is competitive, because it operates on a compact, engineered 1,281-dimensional descriptor rather than raw pixels. As we show earlier in the paper, an SVM directly on pixel intensities performs noticeably worse, so the gain comes from feature engineering, not from the classifier alone. Moreover, the total cost of this purely classical pipeline, feature extraction plus SVM training, remains modest on CPU as both stages are feed-forward, parallelizable, and do not require any backpropagation through a large network.

The utility of the handcrafted descriptor becomes more evident in a hybrid setting. Once deep embeddings from a lightweight CNN (MobileNetV2) are available, concatenating these 1{,}280-dimensional embeddings with our 1{,}281-dimensional handcrafted features and training a linear SVM on the fused 2{,}561-dimensional space not only surpasses the CNN's standalone test accuracy (93.10\% vs.\ 92.43\%), but also requires only a small fraction of the training time compared to end-to-end CNN fine-tuning. In this regime, the costly deep training is performed once to obtain reusable embeddings, and the handcrafted features provide an efficient, CPU-friendly way to further improve accuracy.

\begin{table}[t]
	\centering
	\caption{Computational cost comparison on CIFAR-10 (CPU-only). Feature extraction times are measured over the full dataset (train + validation + test); classifier training times are measured on the train/validation split.}
	\label{tab:cifar_efficiency}
	\begin{tabularx}{\linewidth}{@{}lXr@{}}
		\toprule
		\textbf{Setting} & \textbf{Value} & \textbf{Notes} \\
		\midrule
		\midrule

		\multicolumn{3}{@{}l@{}}{\textbf{MobileNetV2 (end-to-end fine-tuning)}} \\
		\addlinespace
		Feature dim.
		  & 1{,}280
		  & Concatenation \\
		\addlinespace
		Feature ext. time (s)
		  & $\approx 93$
		  & Full dataset \\
		\addlinespace
		Classifier train time (s)
		  & 23{,}054
		  & Train/val split \\
		\addlinespace
		Test acc. (\%)
		  & 92.43\%
		  & CPU-only \\
		\addlinespace
		\addlinespace

		\multicolumn{3}{@{}l@{}}{\textbf{Handcrafted features + SVM}} \\
		\addlinespace
		Feature dim.
		  & 1{,}281
		  & Concatenation \\
		\addlinespace
		Feature ext. time (s)
		  & 898
		  & Full dataset \\
		\addlinespace
		Classifier train time (s)
		  & 4{,}950
		  & Train/val split \\
		\addlinespace
		Test acc. (\%)
		  & 64.81\%
		  & CPU-only \\
		\addlinespace
		\addlinespace

		\multicolumn{3}{@{}l@{}}{\textbf{MobileNetV2 + handcrafted + SVM}} \\
		\addlinespace
		Feature dim.
		  & 2{,}561
		  & Concatenation \\
		\addlinespace
		Feature ext. time (s)
		  & $898 + 93$
		  & Full dataset \\
		\addlinespace
		Classifier train time (s)
		  & 313
		  & Train/val split \\
		\addlinespace
		Test acc. (\%)
		  & 93.10\%
		  & CPU-only \\

		\bottomrule
	\end{tabularx}
\end{table}


Table~\ref{tab:cifar_efficiency} summarizes the key timing and accuracy metrics. Feature extraction times are measured over the full CIFAR-10 dataset (45k train, 5k validation, 10k test), and classifier training times are measured on the train/validation split only. In terms of computation, the handcrafted SVM pipeline offers a competitive classical baseline by using compact 1{,}281-dimensional features with modest CPU-only costs (898\,s feature extraction + 4{,}950\,s training), clearly improving over pixel-based SVMs. Moreover, when fused with MobileNetV2 embeddings, a linear SVM reaches the best accuracy (93.10\% vs.\ 92.43\% for MobileNetV2 alone) while requiring only 313\,s of classifier training compared to 23{,}054\,s for end-to-end CNN fine-tuning, highlighting the time-efficiency of our hybrid approach.

As future work, we plan to explore more systematic hybrids of deep and handcrafted features by varying both the backbone (e.g., other CNNs and ViT variants) and the feature set (e.g., alternative entropy measures and color-space variants of HOG/LBP). It would also be interesting to study automated feature selection or compression on top of the proposed descriptor to further optimize the accuracy–efficiency trade-off and to validate the approach on additional, larger-scale benchmarks.

\section{Conclusion}\label{sec:Conclusion}

To conclude, we presented a unified handcrafted feature extraction pipeline that integrates diverse descriptors, including temporal-complexity measures such as permutation entropy, spatial correlations, edge- and texture-based features (e.g., HOG, LBP), to construct a rich and interpretable feature representation. By applying this approach to a suite of benchmark datasets, including Fashion-MNIST, KMNIST, EMNIST, and CIFAR-10, we demonstrated that shallow classifiers, particularly SVM, can achieve competitive performance without relying on deep learning architectures.

Our results show that combining complementary handcrafted features significantly outperforms traditional single-feature methods in the classical machine learning regime (Fig.~\ref{fig_6}). Notably, we achieve up to \textbf{91.23\% test accuracy} on Fashion-MNIST and \textbf{64.81\%} on CIFAR-10 using only interpretable, resource-efficient techniques. Moreover, beyond serving as a strong classical baseline, the proposed handcrafted descriptors also complement existing deep learning architectures. For example, when fused with MobileNetV2 embeddings, a linear SVM achieves the best overall accuracy (\textbf{93.10\%}) compared to MobileNetV2 alone (\textbf{92.43\%}) on CIFAR-10. This improvement is obtained with substantially lower classifier training cost (313 s for the linear SVM) than end-to-end CNN fine-tuning (23,054 s), while preserving interpretability in the handcrafted component. Overall, these results reinforce that feature engineering remains relevant, not only as an efficient alternative in constrained settings, but also as a practical means to enhance deep models. This suggests hybrid strategies will remain valuable in future vision systems.

Furthermore, in the future we plan to use these features as input representations for designing next-generation neural networks. Instead of feeding raw pixels into a deep model, we could build architectures that learn from more meaningful, structured inputs, like the features we extract here. This could lead to models that are faster to train, more accurate with less data, and easier to interpret. Therefore, combining the strengths of handcrafted features with the flexibility of deep learning could pave the way for next-generation computer vision systems that are both efficient and explainable.


\appendices

\section{FashionMNIST and CIFAR-10 workflow (Algorithm box and parameters)}

\begin{figure}[t]
\centering
\fbox{
\begin{minipage}{0.95\linewidth}
\textbf{Algorithm 1: Fashion-MNIST workflow }\\[2pt]
\hrule
\vspace{2pt}
\hrule
\vspace{0.5cm}

\textbf{Input:} $(X_{\mathrm{train}},y_{\mathrm{train}})$, $(X_{\mathrm{test}},y_{\mathrm{test}})$; PE $(m,\tau)$; patch $(p,s)$; diagonal offsets $\mathcal{K}$; HOG/LBP; SVM grid $\{C,\gamma\}$.\\
\textbf{Output:} tuned RBF-SVM; test accuracy.
\begin{enumerate}
\item Define permutation entropy $PE(\mathbf{t};m,\tau)$ on a 1D sequence $\mathbf{t}$ (normalized by $\log_2(m!)$).
\item For each image $I$:
\begin{enumerate}
\item Row/column directional PE: $\sqrt{PE(\text{forward})\cdot PE(\text{reverse})}$ for all rows/columns.
\item Adjacent-row and adjacent-column Pearson correlations; set undefined values (NaN) to 0.
\item Diagonal and anti-diagonal PE over offsets $k\in\mathcal{K}$ using $\mathrm{diag}(I,k)$ and $\mathrm{diag}(\mathrm{fliplr}(I),k)$.
\item Patch-wise PE using a $p\times p$ window and stride $s$ on vectorized patches.
\item HOG descriptor and uniform-LBP normalized histogram.
\item Concatenate all blocks into a fixed-length vector $\phi(I)$.
\end{enumerate}
\item Stack $\phi(I)$ into $\Phi_{\mathrm{train}}$ and $\Phi_{\mathrm{test}}$.
\item Fit \texttt{StandardScaler} on $\Phi_{\mathrm{train}}$ and transform train/test. 
\item Tune $(C,\gamma)$ via CV over the predefined grid; train final SVM on the full standardized training set.
\item Evaluate accuracy on the standardized test set.
\end{enumerate}
\end{minipage}
}
\end{figure}

\begin{table}[t]
\centering
\caption{Parameters for the Fashion-MNIST pipeline}
\label{tab:fashion_params}
\begin{tabularx}{\textwidth}{l X}
\toprule
\textbf{Component} & \textbf{Setting (value)} \\
\midrule
\midrule
Dataset & Fashion-MNIST: train 60{,}000; test 10{,}000; grayscale $28\times 28$ \\
Feature parallelism & joblib Parallel: n\_jobs = 50 (verbose progress) \\

\midrule
Permutation Entropy (PE) & order $m=3$, delay $\tau=1$; normalized by $\log_2(m!)$ \\
Row/col PE aggregation & $\sqrt{PE(\text{forward})\cdot PE(\text{reverse})}$ per row/column \\
Adjacent correlations & Pearson between adjacent rows/cols; NaN $\rightarrow 0$ \\
Diagonal PE & offsets $\mathcal{K}=\{-10,-9,\dots,10\}$; mean of forward/reverse PE \\
Anti-diagonal PE & same as diagonal after left-right flip of image \\
Patch-wise PE & patch size $p=4$, stride $s=2$; PE on vectorized patch \\

\midrule
HOG & pixels\_per\_cell $(4,4)$; cells\_per\_block $(1,1)$; orientations = 9; feature\_vector=True \\
LBP & uniform LBP: $P=16$, $R=2$; histogram normalized \\
LBP histogram bins & $P+2=18$ bins (implemented via bins $\{0,\dots,P+1\}$) \\

\midrule
Feature dimensionality & Entropy/corr block: $321$; HOG block: $441$; LBP block: $18$ \newline Total per image: $780$ \\

\midrule
Standardization & StandardScaler (fit on train, applied to train/test) \\

\midrule
Classifier & RBF-SVM (SVC); probability=True; random\_state=42 \\
Hyperparameter grid & $C \in \{10,50,100,200\}$; $\gamma \in \{0.01,0.005,0.001\}$ \\
Model selection & HalvingGridSearchCV; CV folds = 3; scoring = accuracy \\
Resource schedule & resource=n\_samples; factor=2; min=$\max(500,0.1N)$; max=$N$; aggressive\_elimination=False \\
Search parallelism & HalvingGridSearchCV: n\_jobs = 50 \\

\midrule
Label noise & Default 0.02, but reported run uses noise\_level=0 (disabled) \\
Artifacts saved & scaler and tuned SVM model (joblib dump) \\
\bottomrule
\end{tabularx}
\end{table}

\clearpage
\begin{figure}[t]
\centering
\fbox{
\begin{minipage}{0.95\linewidth}
\textbf{Algorithm 2: CIFAR-10 RGB workflow }\\[2pt]
\hrule
\vspace{2pt}
\hrule
\vspace{0.5cm}

\textbf{Input:} $(X_{\mathrm{train}},y_{\mathrm{train}})$, $(X_{\mathrm{test}},y_{\mathrm{test}})$; PE $(m,\tau)$; patch $(p,s)$; HOG/LBP settings; SVM grid $\{C,\gamma\}$.\\
\textbf{Output:} tuned RBF-SVM; test accuracy.
\begin{enumerate}
\item Define permutation entropy $PE(\mathbf{t};m,\tau)$ on a 1D sequence $\mathbf{t}$ (normalized by $\log_2(m!)$).
\item For each RGB image $x$:
\begin{enumerate}
\item Split $x$ into channels $(x^{(R)},x^{(G)},x^{(B)})$.
\item For each channel image $I \in \{x^{(R)},x^{(G)},x^{(B)}\}$:
\begin{enumerate}
\item Row/column directional PE: $\sqrt{PE(\text{forward})\cdot PE(\text{reverse})}$ for all rows/columns.
\item Adjacent-row and adjacent-column Pearson correlations; set undefined values (NaN) to 0.
\item Patch-wise PE using a $p\times p$ window and stride $s$ on vectorized patches.
\item Patch-wise PE on the horizontally flipped image $\mathrm{fliplr}(I)$.
\item HOG descriptor and uniform-LBP normalized histogram.
\item Concatenate all blocks into a fixed-length channel vector $\phi(I)$.
\end{enumerate}
\item Concatenate channel vectors to form the final RGB descriptor $\phi(x)=[\phi(x^{(R)}),\phi(x^{(G)}),\phi(x^{(B)})]$.
\end{enumerate}
\item Stack $\phi(x)$ into $\Phi_{\mathrm{train}}$ and $\Phi_{\mathrm{test}}$.
\item Fit \texttt{StandardScaler} on $\Phi_{\mathrm{train}}$ and transform train/test.
\item Tune $(C,\gamma)$ via CV over the predefined grid; train final SVM on the full standardized training set.
\item Evaluate accuracy on the standardized test set.
\end{enumerate}
\end{minipage}
}
\end{figure}

\begin{table}[t]
\centering
\caption{Parameters for the CIFAR-10 RGB pipeline}
\label{tab:cifar_params}
\begin{tabularx}{\textwidth}{l X}
\toprule
\textbf{Component} & \textbf{Setting (value)} \\
\midrule
\midrule
Dataset & CIFAR-10: train 50{,}000; test 10{,}000; RGB $32\times 32\times 3$ \\
Feature parallelism & joblib Parallel: n\_jobs = 50 (verbose progress) \\

\midrule
Permutation Entropy (PE) & order $m=3$, delay $\tau=1$; normalized by $\log_2(m!)$ \\
Row/col PE aggregation & $\sqrt{PE(\text{forward})\cdot PE(\text{reverse})}$ per row/column \\
Adjacent correlations & Pearson between adjacent rows/cols; NaN $\rightarrow 0$ \\
Patch-wise PE & patch size $p=8$, stride $s=4$; PE on vectorized patch \\
Flipped patch PE & same patch-wise PE on $\mathrm{fliplr}(I)$ \\

\midrule
HOG (per channel) & pixels\_per\_cell $(8,8)$; cells\_per\_block $(1,1)$; orientations = 9; feature\_vector=True \\
LBP (per channel) & uniform LBP: $P=8$, $R=1$; histogram normalized \\
LBP histogram bins & $P+2=10$ bins (implemented via bins $\{0,\dots,P+1\}$) \\

\midrule
Feature dimensionality &
Per channel: entropy/corr block $273$; HOG block: $144$; LBP block: $10$ \newline
Total per channel: $427$; Total per image: $1281$ \\

\midrule
Standardization & StandardScaler (fit on train, applied to train/test) \\

\midrule
Classifier & RBF-SVM (SVC); probability=True; random\_state=42 \\
Hyperparameter grid & $C \in \{10,50,100,200\}$; $\gamma \in \{0.01,0.005,0.001\}$ \\
Model selection & HalvingGridSearchCV; CV folds = 3; scoring = accuracy \\
Resource schedule & resource=n\_samples; factor=2; min=$\max(500,0.1N)$; max=$N$ \\
Search parallelism & HalvingGridSearchCV: n\_jobs = 50 \\

\midrule
Label noise & Default 0.02, but reported run uses noise\_level=0 (disabled) \\
Artifacts saved & scaler and tuned SVM model (joblib dump) \\
\bottomrule
\end{tabularx}
\end{table}

\newpage

 

\vspace{-33pt}
\begin{IEEEbiographynophoto}{Abhijit Sen}
holds an M.Sc. in Physics from the prestigious Indian Institute of Technology (IIT) Roorkee, India where he specialized in string theory. He further pursued his academic journey by completing the Aspirantura (Ph.D.-equivalent program) at Novosibirsk State University, Russia. Following his doctoral studies, Dr. Sen expanded his research experience as a postdoctoral scholar at
Tulane University, USA. Alongside his academic pursuits, he has also gained valuable industry experience in India, working at the intersection of Artificial Intelligence and Quantum Computing.
\end{IEEEbiographynophoto}

\vspace{-33pt}
\begin{IEEEbiographynophoto}{Giridas Maiti}
received a Ph.D. degree in Geology from Jadavpur University, India, in 2021. He is currently a Postdoctoral Researcher with the Institute of Applied Geosciences, Karlsruhe Institute of Technology (KIT), Germany. His research interests lie in the computational modeling of geodynamic processes, with a focus on high-resolution 3D thermomechanical simulations of mountain building (e.g., the Himalayas and the Alps) and subduction dynamics. In parallel, he explores cutting-edge applications of machine learning and artificial intelligence for the analysis of large-scale geological data and imagery. 
\end{IEEEbiographynophoto}

\vspace{-33pt}
\begin{IEEEbiographynophoto}{Bikram K. Parida}
received an M.Sc. in Physics with a specialization in astrophysics from Pondicherry Central University, India, in 2019. Since 2023, he has been pursuing a PhD in Information and Communication Engineering from Sun Moon University, South Korea. From 2017 to 2019, he worked as a research assistant at the Indian Institute of Astrophysics, India. Since 2023, He is a Research Assistant with the Artificial Intelligence and Image Processing Laboratory, Sun Moon University, South Korea. His research interests include Artificial Intelligence, theoretical physics, Quantum Computations and Astrophysics. His current research is primarily concentrated on computer vision using deep learning techniques, specifically on 3D reconstruction from a single view image using a Neural Radiance Field. He is also interested in implementing Quantum Machine Learning (QML) models in various domains.
\end{IEEEbiographynophoto}


\vspace{-33pt}
\begin{IEEEbiographynophoto}{Bhanu P. Mishra}
     received an M.C.A. degree from Indira Gandhi National Open University (IGNOU), India, in 2021. Since 2022, he has been working as a Computer Vision Engineer at Cureskin, where he is engaged in applied research and system-level deployment of machine learning models for healthcare diagnostics. His current research focuses on deep learning-based detection of hair baldness, combining advanced image analysis with medical domain expertise. He is particularly interested in interpretable computer vision, real-time inference optimization, and robust system design for mobile-first applications. In parallel, he explores the intersection of artificial intelligence and dermatological imaging, with ongoing work in the development of scalable inference pipelines on Linux-based systems. His broader research interests include medical computer vision, model compression, and edge deployment. He emphasizes the use of modular, extensible, and maintainable software design principles to ensure reproducibility, adaptability, and high-performance integration of machine learning systems into production environments.
\end{IEEEbiographynophoto}

\vspace{-33pt}
\begin{IEEEbiographynophoto}{Mahima Arya}
     received her Ph.D. in Physics from the Indian Institute of Technology, Roorkee, India, in 2018, and completed a Certificate program in Artificial Intelligence from the International University of Applied Sciences, Germany, in 2021. She served as an Assistant Professor at the School of Artificial Intelligence, Amrita Vishwa Vidyapeetham, Coimbatore, India, where she taught fundamentals of AI, computer vision, and NLP. Her research interests include artificial intelligence, machine learning, data analysis, and AI-assisted medical image processing. She has authored over 12 peer-reviewed publications and has experience in implementing AI solutions across academic and industry settings.
\end{IEEEbiographynophoto}

\vspace{-33pt}
\begin{IEEEbiographynophoto}
{Denys I. Bondar}, Associate Professor, joined the Department of Physics and Engineering Physics at Tulane University in 2018. Previously, he was an Associate Research Scholar and Lecturer at Princeton University, where he had been since 2014 after being promoted from a postdoctoral appointment. He earned his Ph.D. in Physics from the University of Waterloo, Canada, in 2011. Dr. Bondar has received several prestigious awards, including the W. M. Keck Foundation Award (2021), the DARPA Young Faculty Award (2019), the Humboldt Research Fellowship for Experienced Researchers (2017), the U.S. Air Force Young Investigator Research Program award (2016), and the Los Alamos Director’s Fellowship (2013, declined). The main topic of his research is quantum control, reservoir computing, and developing physics informed ML techniques.
\end{IEEEbiographynophoto}


\vfill


\clearpage
\begin{thebibliography}{1}
\bibliographystyle{IEEEtran}

\bibitem{bandt2002permutation}
Bandt, C., \& Pompe, B. (2002). Permutation entropy: A natural complexity measure for time series. \textit{Physical Review Letters}, 88(17), 174102.  
\url{https://doi.org/10.1103/PhysRevLett.88.174102}

\bibitem{shannon1948mathematical}
Shannon, C. E. (1948). A mathematical theory of communication. \textit{Bell System Technical Journal}, 27(3), 379--423.  
\url{https://doi.org/10.1002/j.1538-7305.1948.tb01338.x}

\bibitem{renyi1961measures}
R\'enyi, A. (1961). On measures of entropy and information. In \textit{Proceedings of the Fourth Berkeley Symposium on Mathematical Statistics and Probability, Volume 1: Contributions to the Theory of Statistics} (pp. 547--561). University of California Press.

\bibitem{nicolaou2012use}
Nicolaou, N., \& Georgiou, J. (2011). Use of permutation entropy to characterize sleep electroencephalograms. \textit{Clinical EEG and Neuroscience}, 43(3), 199--206.  
\url{https://journals.sagepub.com/doi/10.1177/155005941104200107}

\bibitem{zunino2012permutation}
Zanin,M., Zunino, L., Rosso, O. A., \& Papo, D. (2012). Permutation entropy and its main biomedical and econophysics applications: a review. \textit{Entropy}, 14(8), 1553--1577.  
\url{https://doi.org/10.3390/e14081553}

\bibitem{li2016automatic}
Bandt, C. (2017). A new kind of permutation entropy used to classify sleep stages from invisible EEG microstructure. \textit{Entropy}, 19(5), 197.  
\url{https://doi.org/10.3390/e19050197} 

\bibitem{xiao2025mse_asd}
Xiao, W., \& Jones, M. (2025). Multiscale entropy of resting-state fMRI signals reveals differences in brain complexity in autism. \textit{bioRxiv}.  
\url{https://doi.org/10.1101/2025.05.22.655518}


\bibitem{qin2018sbbscnn}
W.~Qin, J.~Wu, F.~Han, Y.~Yuan, W.~Zhao, B.~Ibragimov, J.~Gu, and L.~Xing, “Superpixel‐based and boundary‐sensitive convolutional neural network for automated liver segmentation,” \textit{Phys. Med. Biol.}, vol.~63, no.~9, p. 095017, May 2018.  
\url{https://doi.org/10.1088/1361-6560/aabd19}

\bibitem{kapur1985_entropy_thresholding}
Kapur, J. N., Sahoo, P. K., \& Wong, A. K. C. (1985). A new method for gray-level picture thresholding using the entropy of the histogram. \textit{Computer Vision, Graphics, and Image Processing}, 29, 273--285.  
\url{https://doi.org/10.1016/0734-189X(85)90125-2}

\bibitem{gaudencio2022_pe2d_aape2d}
Gaudêncio, A. S., Hilal, M., Cardoso, J. M., Humeau-Heurtier, A., \& Vaz, P. G. (2022). Texture analysis using two-dimensional permutation entropy and amplitude-aware permutation entropy. \textit{Pattern Recognition Letters}, 159, 150--156.  
\url{https://doi.org/10.1016/j.patrec.2022.05.017}

\bibitem{zunino2016_ms_cecp_textures}
Zunino, L., \& Ribeiro, H. V. (2016). Discriminating image textures with the multiscale two-dimensional complexity-entropy causality plane. \textit{Chaos, Solitons \& Fractals}, 91, 679--688.  
\url{https://doi.org/10.1016/j.chaos.2016.09.005}

\bibitem{morel2021_mpe2d}
Morel, C., \& Humeau-Heurtier, A. (2021). Multiscale permutation entropy for two-dimensional patterns. \textit{Pattern Recognition Letters}, 150, 139--146.  
\url{https://doi.org/10.1016/j.patrec.2021.06.028}


\bibitem{simonyan2015vgg}
Simonyan, K., \& Zisserman, A. (2015). Very deep convolutional networks for large-scale image recognition. In \textit{International Conference on Learning Representations (ICLR)}, San Diego, USA.  
\url{https://arxiv.org/abs/1409.1556}

\bibitem{krizhevsky2009learning}
Krizhevsky, A. (2009). Learning Multiple Layers of Features from Tiny Images. Technical Report TR-2009, University of Toronto.  
\url{https://www.cs.toronto.edu/~kriz/learning-features-2009-TR.pdf}

\bibitem{he2016deep}
He, K., Zhang, X., Ren, S., \& Sun, J. (2016). Deep residual learning for image recognition. In \textit{Proceedings of the IEEE Conference on Computer Vision and Pattern Recognition (CVPR)} (pp. 770--778).  
\url{https://doi.org/10.1109/CVPR.2016.90}

\bibitem{howard2017mobilenets}
Howard, A. G., Zhu, M., Chen, B., Kalenichenko, D., Wang, W., Weyand, T., Andreetto, M., \& Adam, H. (2017). MobileNets: Efficient convolutional neural networks for mobile vision applications. \textit{arXiv preprint arXiv:1704.04861}.  
\url{https://arxiv.org/abs/1704.04861}

\bibitem{xiao2017fashion}
Xiao, H., Rasul, K., \& Vollgraf, R. (2017). Fashion-MNIST: A Novel Image Dataset for Benchmarking Machine Learning Algorithms. \textit{arXiv preprint arXiv:1708.07747}.  
\url{https://arxiv.org/abs/1708.07747}

\bibitem{dalal2005}
Dalal, N., \& Triggs, B. (2005). Histograms of oriented gradients for human detection. In \textit{IEEE Computer Society Conference on Computer Vision and Pattern Recognition (CVPR)} (Vol. 1, pp. 886--893). IEEE.  
\url{https://doi.org/10.1109/CVPR.2005.177}

\bibitem{ojala2002}
Ojala, T., Pietikäinen, M., \& Mäenpää, T. (2002). Multiresolution gray-scale and rotation invariant texture classification with local binary patterns. \textit{IEEE Transactions on Pattern Analysis and Machine Intelligence}, 24(7), 971--987.  
\url{https://doi.org/10.1109/TPAMI.2002.1017623}

\bibitem{cortes1995support}
Cortes, C., \& Vapnik, V. (1995). Support-vector networks. \textit{Machine Learning}, 20(3), 273--297.  
\url{https://doi.org/10.1007/BF00994018}

\bibitem{fashionmnistweb}
Fashion-MNIST Dataset, “Fashion-MNIST: A Novel Image Dataset for Benchmarking Machine Learning Algorithms,” [Online]. Available:  
\url{http://fashion-mnist.s3-website.eu-central-1.amazonaws.com/}

\bibitem{vo2015tensorhog}
Vo, T., Tran, D., \& Ma, W. (2015). Tensor decomposition and application in image classification with histogram of oriented gradients. \textit{Neurocomputing}, 165, 38--45.  
\url{https://doi.org/10.1016/j.neucom.2014.06.093}

\bibitem{zhu2023nonlinearstmhog}
Zhu, C., Zhao, W., \& Lian, H. (2023). Image recognition and classification with HOG based on nonlinear support tensor machine. \textit{Multimedia Tools and Applications}, 82, 20119--20138.  
\url{https://doi.org/10.1007/s11042-022-14320-x}



\bibitem{ranzato2019robustness}
F. Ranzato and M. Zanella, “Robustness Verification of Support Vector Machines,” \textit{CoRR}, vol. abs/1904.11803, 2019.  
\url{https://arxiv.org/abs/1904.11803}


\bibitem{sharifnejad2021fer}
M.~Sharifnejad, A.~Shahbahrami, A.~Akoushideh, and R.~Z.~Hassanpour,  
“Facial expression recognition using a combination of enhanced local binary pattern and pyramid histogram of oriented gradients features extraction,”  
\textit{IET Image Process.}, vol.~15, no.~2, pp. 468–478, Feb. 2021.  
\url{https://doi.org/10.1049/ipr2.12037}


\bibitem{clanuwat2018deep}
Clanuwat, T., Boonma, M., Kachitvichyanukul, D., \& Rungger, A. (2018). Deep learning for classical Japanese literature. \textit{arXiv preprint arXiv:1812.01718}.  
\url{https://arxiv.org/abs/1812.01718}


\bibitem{rois2020svm_kmnist}
“rois-codh/kmnist: Repository for Kuzushiji-MNIST,” GitHub, 2018. [Online]. Available: \url{https://github.com/rois-codh/kmnist} 


\bibitem{cohen2017emnist}
Cohen, G., Afshar, S., Tapson, J., \& van Schaik, A. (2017). EMNIST: Extending MNIST to handwritten letters. \textit{arXiv preprint arXiv:1702.05373}.  
\url{https://arxiv.org/abs/1702.05373}

\bibitem{ghadekar2018dwt_dct}
P.~Ghadekar, S.~Ingole, and D.~Sonone, “Handwritten digit and letter recognition using hybrid DWT–DCT with KNN and SVM classifier,” in \textit{Proc. 2018 4th Int. Conf. Comput. Commun. Control Autom. (ICCUBEA)}, Pune, India, Jul. 2018, pp. 1–6, doi:\url{https://ieeexplore.ieee.org/document/8697684}.


\bibitem{osamakhaan2022cifar_svm_pca}
O.~Khaan, “CIFAR-10 Image Classification using Support Vector Machines and PCA,” GitHub repository, 2022.  
\url{https://github.com/osamakhaan/CIFAR-10-Image-Classification} 

\bibitem{le2013fastfood}
Q.~V. Le, T.~Sarl\'os, and A.~J. Smola, “Fastfood: Approximate kernel expansions in log-linear time,” in \textit{Proc. 30th Int. Conf. Mach. Learn. (ICML)}, Atlanta, GA, USA, Jun. 2013, vol.~28, pp. 244–252.  
[Online]. Available: \url{https://proceedings.mlr.press/v28/le13.pdf}

\bibitem{le2019cifar_svm_ensemble}
T.~V. Le, “Classifying CIFAR-10 images using unsupervised feature learning and ensemble methods,” Tech. Rep., Mar. 2019.  
\url{https://trucvietle.me/files/601-report.pdf}

\bibitem{lemuel2021cifar_hog_pca_svm}
L.K. Lee “CIFAR10‐HOG‐PCA‐SVM: Grayscaling, HOG, PCA and RBF SVM classifier for CIFAR-10,” GitHub repository, 2021.  
\url{https://github.com/LemuelKL/CIFAR10-HOG-PCA-SVM}

\bibitem{coates2011analysis}
Coates, A., Lee, H., \& Ng, A. Y. (2011). An Analysis of Single-Layer Networks in Unsupervised Feature Learning. In \textit{Proceedings of the Fourteenth International Conference on Artificial Intelligence and Statistics} (AISTATS), 15, 215--223.  
\url{https://proceedings.mlr.press/v15/coates11a.html}

\bibitem{HOG1}
Sun, Z., Santos, J., \& Caetano, E. (2022). Vision and support vector machine--based train classification using weigh-in-motion data. \textit{Journal of Bridge Engineering}, 27, 06022001.  
\url{https://doi.org/10.1061/(ASCE)BE.1943-5592.0001878}

\bibitem{HOG2}
Xu, Y., Imou, K., Kaizu, Y., \& Saga, K. (2013). Two-stage approach for detecting slightly overlapping strawberries using HOG descriptor. \textit{Biosystems Engineering}, 115, 144--153.  
\url{https://doi.org/10.1016/j.biosystemseng.2013.03.011}





\end{thebibliography}
\end{document}